\documentclass[journal]{IEEEtran}
\usepackage{cite}
\usepackage{hyperref}
\usepackage{tabularx}
\usepackage{xspace}
\makeatletter
\DeclareRobustCommand\onedot{\futurelet\@let@token\@onedot}
\def\@onedot{\ifx\@let@token.\else.\null\fi\xspace}

\makeatother

\ifCLASSINFOpdf
   \usepackage[pdftex]{graphicx}
\else
   \usepackage[dvips]{graphicx}
\fi
\usepackage{makecell}
\usepackage{booktabs}
\usepackage{xcolor}
\usepackage{multirow}

\usepackage{amsmath,amssymb}

\usepackage{pifont}

\usepackage{url}
\begin{document}
%
% paper title
% Titles are generally capitalized except for words such as a, an, and, as,
% at, but, by, for, in, nor, of, on, or, the, to and up, which are usually
% not capitalized unless they are the first or last word of the title.
% Linebreaks \\ can be used within to get better formatting as desired.
% Do not put math or special symbols in the title.
\title{Multi-Modal Motion Retrieval by Learning a Fine-Grained Joint Embedding Space}
%
%
% author names and IEEE memberships
% note positions of commas and nonbreaking spaces ( ~ ) LaTeX will not break
% a structure at a ~ so this keeps an author's name from being broken across
% two lines.
% use \thanks{} to gain access to the first footnote area
% a separate \thanks must be used for each paragraph as LaTeX2e's \thanks
% was not built to handle multiple paragraphs

\author{Shiyao Yu, Zi-An Wang, Kangning Yin, Zheng Tian, Mingyuan Zhang, Weixin Si and Shihao Zou
\thanks{S. Yu is with Southern University of Science and Technology, and jointly with Shenzhen Institutes of Advanced Technology, Chinese Academy of Sciences, Shenzhen, China. (E-mail: sy.yu1@siat.ac.cn).}
\thanks{Z. Wang is with University of Chinese Academy of Sciences, and jointly with Shenzhen Institutes of Advanced Technology, Chinese Academy of Sciences, Shenzhen, China. (E-mail: za.wang@siat.ac.cn).}
\thanks{K. Yin and Z. Tian are with ShanghaiTech University, Shanghai, China. (E-mail: yinkn2022@shanghaitech.edu.cn, tianzheng@shanghaitech.edu.cn).}
\thanks{M. Zhang is with Nanyang Technological University, Singapore. (E-mail: mingyuan001@e.ntu.edu.sg).}
\thanks{W. Si is with Faculty of Computer Science and Control Engineering, Shenzhen University of Advanced Technology, Shenzhen, China. (E-mail: wx.si@siat.ac.cn).}
\thanks{S. Zou is with Shenzhen Institutes of Advanced Technology, Chinese Academy of Sciences, Shenzhen, China. (E-mail: sh.zou@siat.ac.cn).}
\thanks{Shihao Zou is the corresponding author.}
% <-this % stops a space
% \thanks{Manuscript received January 05 2022; accepted March 18 2022.}
}
% Shihao Zou is the corresponding author for this paper.

% note the % following the last \IEEEmembership and also \thanks - 
% these prevent an unwanted space from occurring between the last author name
% and the end of the author line. i.e., if you had this:
% 
% \author{....lastname \thanks{...} \thanks{...} }
%                     ^------------^------------^----Do not want these spaces!
%
% a space would be appended to the last name and could cause every name on that
% line to be shifted left slightly. This is one of those "LaTeX things". For
% instance, "\textbf{A} \textbf{B}" will typeset as "A B" not "AB". To get
% "AB" then you have to do: "\textbf{A}\textbf{B}"
% \thanks is no different in this regard, so shield the last } of each \thanks
% that ends a line with a % and do not let a space in before the next \thanks.
% Spaces after \IEEEmembership other than the last one are OK (and needed) as
% you are supposed to have spaces between the names. For what it is worth,
% this is a minor point as most people would not even notice if the said evil
% space somehow managed to creep in.

% The paper headers
\markboth{Journal of \LaTeX\ Class Files,~Vol.~14, No.~8, August~2015}%
{Shell \MakeLowercase{\textit{et al.}}: Bare Demo of IEEEtran.cls for IEEE Journals}
% The only time the second header will appear is for the odd numbered pages
% after the title page when using the twoside option.
% 
% *** Note that you probably will NOT want to include the author's ***
% *** name in the headers of peer review papers.                   ***
% You can use \ifCLASSOPTIONpeerreview for conditional compilation here if
% you desire.

% If you want to put a publisher's ID mark on the page you can do it like
% this:
%\IEEEpubid{0000--0000/00\$00.00~\copyright~2015 IEEE}
% Remember, if you use this you must call \IEEEpubidadjcol in the second
% column for its text to clear the IEEEpubid mark.

% use for special paper notices
%\IEEEspecialpapernotice{(Invited Paper)}

% make the title area
\maketitle

% As a general rule, do not put math, special symbols or citations
% in the abstract or keywords.
\begin{abstract}
Motion retrieval is crucial for motion acquisition, offering superior precision, realism, controllability, and editability compared to motion generation. Existing approaches leverage contrastive learning to construct a unified embedding space for motion retrieval from text or visual modality. However, these methods lack a more intuitive and user-friendly interaction mode and often overlook the sequential representation of most modalities for improved retrieval performance. 
To address these limitations, we propose a framework that aligns four modalities—text, audio, video, and motion—within a fine-grained joint embedding space, incorporating audio for the first time in motion retrieval to enhance user immersion and convenience. This fine-grained space is achieved through a sequence-level contrastive learning approach, which captures critical details across modalities for better alignment.
To evaluate our framework, we augment existing text-motion datasets with synthetic but diverse audio recordings, creating two multi-modal motion retrieval datasets. Experimental results demonstrate superior performance over state-of-the-art methods across multiple sub-tasks, including an 10.16\% improvement in R@10 for text-to-motion retrieval and a 25.43\% improvement in R@1 for video-to-motion retrieval on the HumanML3D dataset. Furthermore, our results show that our 4-modal framework significantly outperforms its 3-modal counterpart, underscoring the potential of multi-modal motion retrieval for advancing motion acquisition.
% that audio performs comparably to text in motion retrieval and
\end{abstract}

% Note that keywords are not normally used for peerreview papers.

% For peer review papers, you can put extra information on the cover
% page as needed:
% \ifCLASSOPTIONpeerreview
% \begin{center} \bfseries EDICS Category: 3-BBND \end{center}
% \fi
%
% For peerreview papers, this IEEEtran command inserts a page break and
% creates the second title. It will be ignored for other modes.
\IEEEpeerreviewmaketitle

\section{Introduction}
% Recently, generative models and multi-modality models have developed rapidly, showing significant potential in human motion generation~\cite{zhu2023human}. By integrating multi-modality data, generation model leverages complementary information across modalities to produce the required motion sequences. While prior works have achieved significant success, several challenges remain.

% background, connect the topic (human motion acquisition) with the journal topic -> multi-modal conditioned human motion generation -> its drawbacks -> motion retrieval -> current work does not support multi-modal conditioned retrieval 
% This research is especially relevant in the development of interactive media where realistic character movements are essential. 
% Human motion acquisition is a key focus in multimedia research, with significant implications across various domains such as animation, gaming, filmmaking, and robotics.~\cite{koksal2023controllable, zhao2024ta2v} By applying real-world human movements, digital characters can be rendered with more natural, lifelike motion, enhancing user experience and visual realism. Basically, there are two primary approaches to achieving this: retrieving human motions from existing datasets or generating new motions through algorithms~\cite{yan2018structure}. 

Human motion acquisition has attracted increasing research attentions in recent years within multimedia research due to its broad applications in areas such as animation, gaming, filmmaking, robotics and virtual/augmented reality~\cite{wu2025dual,ji2025oneshot,zhao2024ta2v,wang2024crossmodal}. By incorporating real-world human movements, digital characters can exhibit more natural and lifelike motion, enhancing both visual realism and user experience. Generally, human motion can be obtained through two primary approaches: retrieving motions from existing datasets or generating new ones using algorithmic techniques.

Recent advancements in generative modeling have introduced various approaches for synthesizing motion sequences based on multi-modal conditions, such as action~\cite{degardin2022generative,petrovich2021action,lee2023multiact,chen2023executing}, text~\cite{plappert2016kit,guo2022generating,tevet2023human,zhang2023generating}, music~\cite{wang2022groupdancer,le2023music,tseng2023edge,gao2022pc}, or scenes~\cite{zhang2022couch,taheri2022goal}. These approaches reduce human effort and eliminate the need for costly motion capture systems. However, a major limitation of generative models is their black-box nature, which restricts control over the generation process, often resulting in unpredictable and difficult-to-modify outputs~\cite{zhu2023human}. Consequently, multi-modal motion generation frequently falls short of expectations, particularly in complex real-world scenarios.

In contrast, motion retrieval overcomes the limitations of generative models by enabling the selection of specific human motion sequences from large datasets. This approach is particularly valuable in industries where precision, realism, controllability, and editability are essential. Early research in this field~\cite{holden2017phase,holden2020learned} primarily focused on retrieving natural motion based on user control signals but did not incorporate the multi-modal conditions explored in recent motion generation studies. Recent advancements in multi-modal motion retrieval, such as TMR~\cite{petrovich2023tmr} and LAVIMO~\cite{yin2024tri}, leverage contrastive learning to construct a unified embedding space for multiple modalities, enabling motion retrieval from diverse inputs such as video and text descriptions. This shift toward multi-modal integration marks a significant step in enhancing the versatility and applicability of motion retrieval in complex scenarios.

\begin{figure}[t]
  \centering
  \includegraphics[width=\columnwidth]{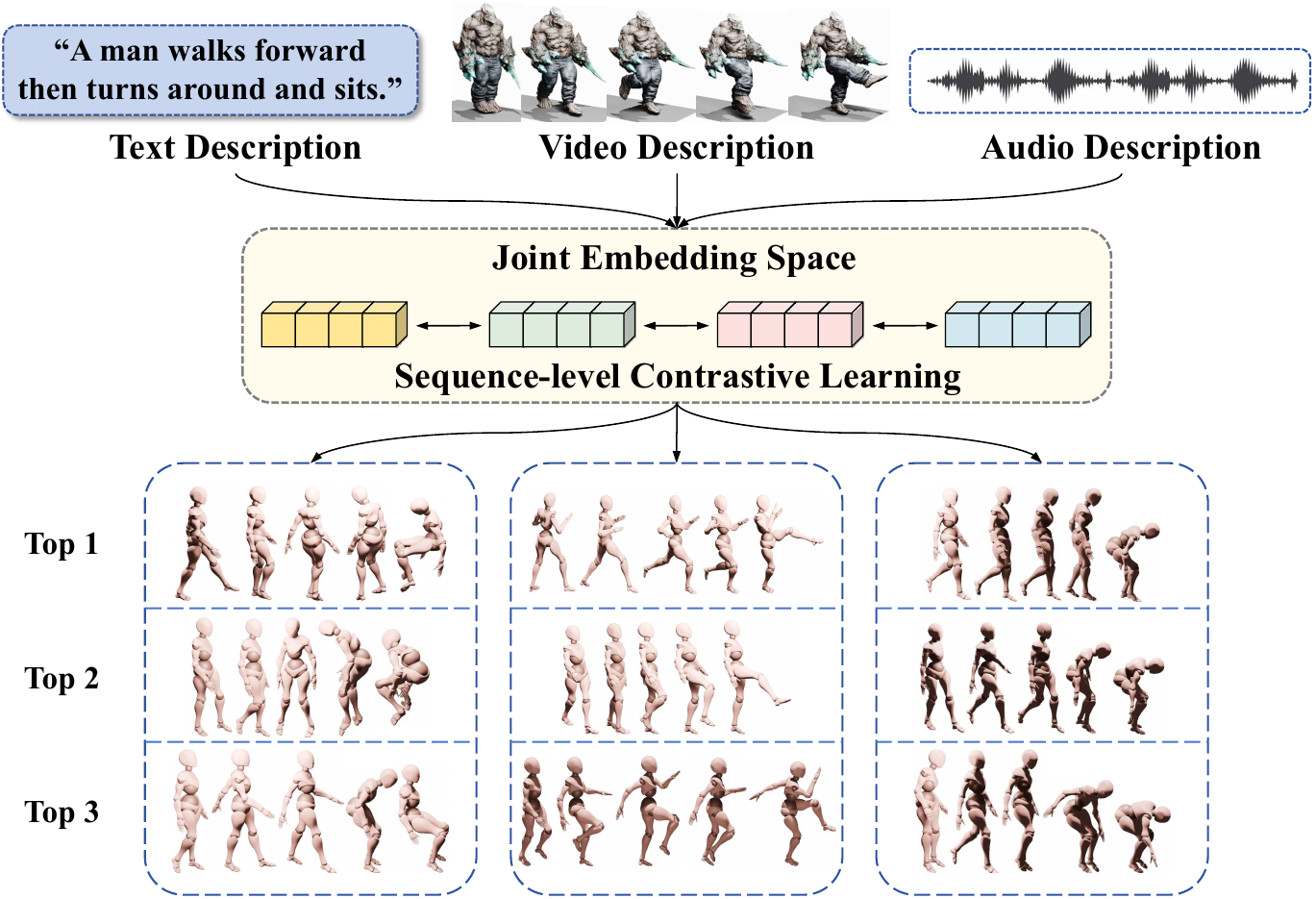}
  \caption{
    \textbf{Overview of Our Work.} Our framework encodes text, video, or audio descriptions and computes their similarity within a shared joint embedding space, ranking candidate motions based on similarity scores to retrieve the most relevant motion.
    }
    \vspace{-4mm}
    \label{fig:introduction}
\end{figure}

\begin{figure*}[h]
  \centering
  \includegraphics[width=0.9\textwidth]{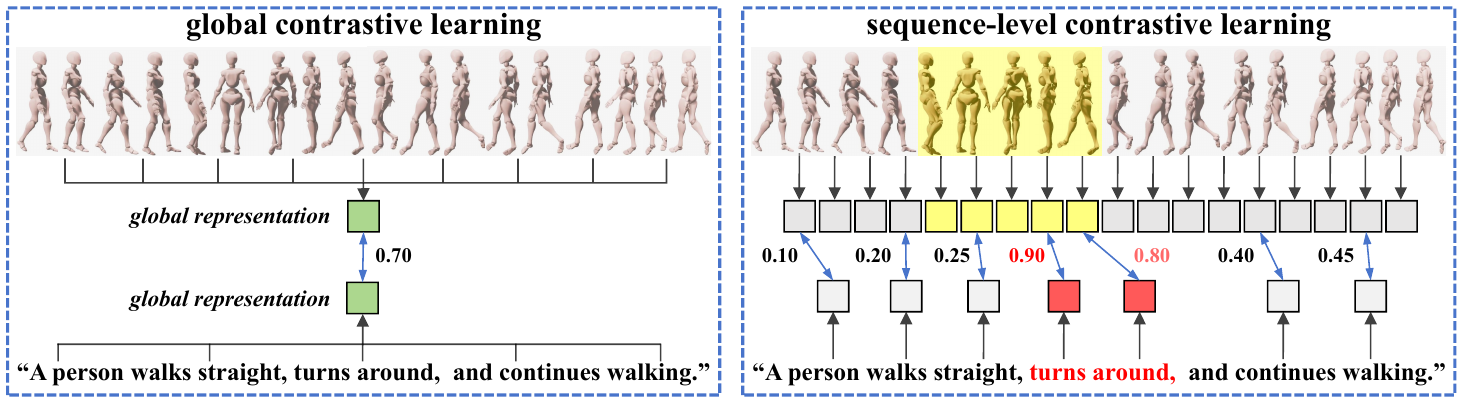}
  \caption{
     \textbf{Global contrastive learning} (Left) computes similarity between two modalities using global representations, where motion and text data are compressed into a single token for cross-modal alignment. In contrast, \textbf{sequence-level contrastive learning} (Right) aligns individual tokens with their most relevant counterparts, enabling the model to focus on key frames in the motion sequence and important keywords in the text description (highlighted in \textcolor{yellow}{yellow} and \textcolor{red}{red}, respectively). As illustrated by the example, the phrase “turns around” yields higher similarity scores with the corresponding frames in the motion sequence, thereby enabling more accurate alignment between the text and motion pair.
    }
    \label{fig:seq-level align}
\end{figure*}

However, previous works~\cite{petrovich2023tmr,yin2024tri} primarily focus on text or visual modalities, overlooking practical challenges in real-world applications. These modalities often require users to provide detailed and specific references, which can be time-consuming and unintuitive. Consequently, they are less suited for environments that demand natural and fluid interactions to sustain user engagement. In contrast, audio-based instructions provide a more immediate and intuitive communication channel, enabling seamless semantic understanding and enhancing the interaction experience. Recent advancements in interactive technologies, such as GPT-4o~\cite{achiam2023gpt}, further underscore the potential of audio as a modality, significantly improving user immersion and convenience.

% Additionally, the joint embedding space for motion retrieval is learned through global alignment in prior works~\cite{petrovich2023tmr, yin2024tri}, where sequential features from each modality are reduced to a single vector and aligned with other modalities using contrastive learning. However, many modalities, such as video, text, and audio, are sequential, and key information may only be present in a small segment of the sequence. For instance, as illustrated in Fig.~\ref{fig:seq-level align}, the phrase “turns around” in a lengthy text corresponds to just a few frames in the middle of the motion sequence. The global alignment approach tends to mix up these critical details, leading to suboptimal alignment and poorer retrieval performance.
% Furthermore, the joint embedding space for motion retrieval was learned through global alignment in prior works~\cite{petrovich2023tmr,yin2024tri}, where sequential features from each modality are compressed into a single vector and aligned using contrastive learning. However, many modalities—such as video, text, and audio—are inherently sequential, with key information often confined to specific segments of the sequence. As illustrated in the example of Fig.~\ref{fig:seq-level align}, the phrase “turns around” within a long text description corresponds to only a few frames in the middle of the motion sequence. Global alignment methods tend to overlook these crucial details, leading to suboptimal alignment and diminished retrieval performance.
Furthermore, previous works~\cite{petrovich2023tmr,yin2024tri} learned the joint embedding space for motion retrieval through global alignment, where sequential features from each modality are compressed into a single vector and aligned using contrastive learning. However, many modalities—such as video, text, and audio—are inherently sequential, with key information often concentrated in specific segments. For instance, as shown in Fig.~\ref{fig:seq-level align}, the phrase “turns around” within a lengthy text description corresponds to only a few frames in the middle of the motion sequence. Global alignment methods tend to overlook such crucial details, resulting in suboptimal alignment and reduced retrieval performance.

To address these challenges, we propose a framework that aligns four modalities—text, audio, video, and motion—within a fine-grained joint embedding space, introducing audio into multi-modal motion retrieval for the first time. This fine-grained embedding is learned through sequence-level alignment, which preserves critical details across modalities by selecting the maximum similarity score when comparing each token with all tokens from another modality.

% To further enhance motion retrieval, we decompose the pose representation into individual body parts before feeding it into the motion encoder. This approach produces a more detailed and nuanced representation of motion tokens, improving alignment accuracy in the subsequent modality-matching stage. Meanwhile, raw audio signals are often sparse and lengthy, making them challenging to encode effectively. We propose to use WavLM~\cite{chen2022wavlm} for feature extraction and a memory-retrieval-based attention module that condenses these features into more compact, model-friendly representations.
To further enhance motion retrieval, we decompose the pose representation into individual body parts before feeding it into the motion encoder. This decomposition provides a more detailed and nuanced representation of motion tokens, improving alignment accuracy in the subsequent modality-matching stage. Meanwhile, raw audio signals are often sparse and lengthy, posing challenges for effective encoding. To address this, we utilize WavLM~\cite{chen2022wavlm} for feature extraction and introduce a memory-retrieval-based attention module that condenses these features into more compact, model-friendly representations. As no existing human motion dataset is explicitly paired with audio instructions, we propose augmenting text-motion datasets by synthesizing audio from text. To enhance audio diversity, we use ChatGPT~\cite{achiam2023gpt} to rephrase textual descriptions into a more conversational, spoken-language style before synthesizing audio recordings with various speaker identities using Tortoise~\cite{betker2023better}.

Our main contributions are summarized as follows: (1) We propose a joint embedding space that aligns four modalities—text, audio, video, and motion—enabling versatile tasks such as text-motion, audio-motion, and video-motion retrieval. (2) We introduce a fine-grained contrastive learning mechanism that computes token-level similarities between sequences, improving alignment accuracy and creating a more precise joint embedding space across all four modalities. (3) We provide two multi-modal motion retrieval datasets by augmenting existing text-motion datasets—KIT-ML~\cite{petrovich2023tmr} and HumanML3D~\cite{guo2022generating}—with conversational audio instructions featuring diverse speaker identities. (4) Our framework achieves state-of-the-art performance in motion retrieval across both datasets, outperforming previous methods by 10.16\% R@10 in text-to-motion and 25.43\% R@1 in video-to-motion on HumanML3D dataset. Additionally, our method demonstrates competitive performance in text-to-motion and audio-to-motion retrieval, highlighting the effectiveness of audio as a retrieval modality.

% (4) Our framework achieves superior results in text-to-motion retrieval on the HumanML3D dataset, outperforming the current state-of-the-art by 3.04\%, 2.97\%, and 3.14\% in R@1, R@5, and R@10, respectively.\shihao{text-to-motion use one number, video-to-motion use one number, audio-to-motion use one number to highlight that audio presents the same utility with text}

% We achieve superior performance in the task of multi-modal motion retrieval, with improvements of 3.04\%, 2.97\%, and 3.14\% in R@1, R@5, and R@10, respectively, over the SOTA method.

% We introduce an encoder that specifically focuses on the spatial information within motion data. This encoder separates the spatial information of the motion, allowing the model to attend to this information during alignment. This approach not only enhances retrieval accuracy but also paves the way for more expansive and deeper multi-modality human motion research.

Our preliminary work was published in~\cite{yin2024tri}. This paper extends our preceding effort in a number of aspects: 
\begin{itemize}
    \item Extending the original three-modal framework, we incorporate audio as an additional modality, enabling richer and more natural user interactions.
    \item Expanding upon existing datasets, we augment with conversational audio instructions, thereby constructing more comprehensive multi-modal human motion datasets.
    \item Departing from global alignment, we introduce sequence-level alignment across different modalities, refining the joint embedding space and significantly improving retrieval performance.
    \item Our four-modal framework outperforms previous two or three-modal approaches across multiple retrieval tasks, including text-motion, video-motion and audio-motion retrieval.
    
\end{itemize}

\section{Related Work}
\subsection{Motion Generation and Motion Retrieval}

% Existing motion generation methods can be categorized into two classes. The first category of approaches relies on regression models that attempt to directly map the input features to the corresponding motion sequence. A representative work of this approach is MotionCLIP~\cite{tevet2022motionclip}. 
% Another category of approaches relies on generative models which seek to model the distribution of motion (or the joint distribution that includes input conditions) in an unsupervised learning framework. Common generative approaches encompass models such as Variational Autoencoders~\cite{kingma2013auto} and Diffusion processes~\cite{ho2020denoising}. Representative works in this category include TEMOS~\cite{petrovich2022temos}, TM2T~\cite{guo2022tm2t}, T2M-GPT~\cite{zhang2023generating}, and MDM~\cite{tevet2023human}. 
% It also incorporates VQ-VAE, along with an Exponential Moving Average and a code reset strategy.

There are two primary approaches to acquiring human motions: one is generating new motions through algorithms, and the other is retrieving related motions from existing datasets. Motion generation has become a popular research area in recent years. MotionCLIP~\cite{tevet2022motionclip} uses a transformer architecture to directly map text inputs to motions. By learning the joint distribution of motion and text through VAE~\cite{kingma2013auto}, TEMOS~\cite{petrovich2022temos} enables the generation of diverse motion sequences. TM2T~\cite{guo2022tm2t} utilizes VQ-VAE~\cite{van2017neural} to jointly train the text-to-motion and motion-to-text modules. Likewise, T2M-GPT~\cite{zhang2023generating} employs a GPT-like transformer architecture for motion sequence generation. Instead of predicting only noise, MDM~\cite{tevet2023human} uses a diffusion model to generate samples at each step. However, the black-box nature of generative models limits control over the generation process, often resulting in unpredictable and difficult-to-modify outputs, particularly in complex real-world scenarios.

% These approaches collectively advance motion generation by leveraging different modeling strategies to improve accuracy, diversity, and flexibility.

Motion retrieval has attracted increasing attention as a complementary approach to obtain motion sequences. Inspired by image-text models such as BLIP~\cite{li2022blip} and CoCa~\cite{yucoca}, TMR~\cite{petrovich2023tmr} pioneered the field of text-motion retrieval, with the primary goal of retrieving corresponding motion sequences from a dataset based on textual descriptions. By constructing a cross-modality joint embedding space, TMR enables text descriptions to retrieve the most semantically relevant motion sequences based on similarity computation. LAVIMO~\cite{yin2024tri}  model extended the TMR by introducing an additional video modality to assist in aligning text and motion, significantly improving retrieval performance. Recent advances in fine-grained image retrieval (FGIR), such as DAHNet~\cite{jiang2024global} and DVF~\cite{jiang2024dvf}, have demonstrated the effectiveness of leveraging both global and local semantic cues to learn compact yet discriminative representations. In particular, DVF emphasizes subcategory-specific discrepancies by combining object-oriented and semantic-oriented visual filtering modules. Inspired by these findings, we explore the use of local information in the context of motion retrieval to improve retrieval accuracy. To this end, our model incorporates a fine-grained contrastive learning mechanism and introduces the audio modality to enhance text-motion alignment. The inclusion of the audio modality also improves user interaction and experience in practical applications.

% To further explore the potential of this framework, our model incorporates a fine-grained contrastive learning mechanism and introduces the audio modality to enhance text-motion alignment. The addition of the audio modality also improves the user experience in practical applications.

\subsection{Foundation Model}

The text encoder usually refers to the part responsible for creating the semantic embedding of each word. Built on the transformer, BERT~\cite{devlin2018bert} processes text bidirectionally, enabling it to understand context from both forward and backward directions. This bidirectional capability significantly enhances its ability to capture semantic relationships. However, this advantage comes at a cost. Despite its performance, BERT's large model size and high computational cost pose challenges for practical deployment, especially in resource-constrained environments. To address this, DistilBERT~\cite{sanh2019distilbert} was developed using knowledge distillation. By transferring knowledge from a larger teacher model to a smaller student model, DistilBERT reduces the number of layers and parameters, achieving faster inference and lower memory usage while maintaining most of BERT's performance. Unlike BERT, which is limited to classification and span prediction, T5~\cite{raffel2020exploring} reformulates all NLP tasks into a text-to-text framework, where both input and output are text strings. This unified approach enables T5 to handle diverse tasks such as translation, summarization, and question answering with the same model and loss function. LLaMA~\cite{touvron2023llama} is a foundational language model designed to support research by providing smaller, more efficient models for studying large language models. Its reduced computational requirements make it a valuable tool for testing new approaches, validating research, and exploring novel applications.

The audio encoder typically refers to the component responsible for extracting the semantic features from audio signals. There have been significant advances in large audio models, revolutionizing the processing of audio across diverse applications. Encodec~\cite{borsos2023audiolm} effectively captures the characteristics of the original audio, while also improving the details of recovered speech and music. This is achieved by integrating the reconstruction loss with the use of discriminator mechanisms. Likewise, Whisper~\cite{radford2023robust} uses an encoder-decoder architecture to extract features from the Log-Mel spectrogram for speech recognition. It achieves impressive multilingual capabilities, made possible by the vast magnitude of training data it is built upon. However, Encodec and Whisper tend to introduce biases in the latent representations of audio signals, which restrict their generalizability. In contrast, WavLM~\cite{chen2022wavlm} leverages a self-supervised masked speech denoising framework, learning universal speech representations from extensive unlabeled datasets. This enables WavLM to effectively adapt to a wide range of speech processing tasks. WavLM excels in 13 speech tasks under the SUPERB~\cite{yang2021superb} benchmark, including speaker recognition, speech recognition, keyword spotting, emotion recognition, and speech enhancement, showing that WavLM is a versatile tool for a wide range of downstream speech processing tasks.

% Additionally, ImageBind takes an innovative approach to multi-modal alignment by utilizing image-paired data, enabling the creation of a shared representation space for audio, images, and videos. This approach significantly improves retrieval and classification tasks, particularly in zero-shot and few-shot settings. These developments enable higher fidelity, greater efficiency, and enhanced versatility in processing audio data.

The video encoder generally refers to the part responsible for extracting the semantic features from video frames. ResNet leverages deep residual learning to address the vanishing gradient problem, enabling the training of very deep networks for image classification tasks. Similarly, ViT~\cite{dosovitskiy2020image} employs a transformer-based architecture for image recognition, treating images as sequences of patches to capture long-range dependencies and spatial relationships. CLIP~\cite{radford2021learning} is a powerful multi-modal model that establishes connections between images and text. CLIP consists of an image encoder and a text encoder. The image encoder typically uses Vision Transformers or ResNet to extract image features, providing flexibility in balancing performance and efficiency. Both encoders share a common vector space, where matching image and text vectors are brought closer, and non-matching vectors are pushed apart, allowing the model to effectively align images and text. VideoPoet~\cite{kondratyuk2023videopoet} is a model that generates diverse video content by leveraging a transformer-based architecture to capture both spatial and temporal dependencies, enabling the synthesis of video sequences from textual descriptions. Recent work such as Label Independent Memory (LIM)~\cite{zhu2020label} leverages a memory module to store generalized features from unlabeled videos, enabling robust prototype construction for few-shot classification. Additionally, THAN~\cite{li2023memory} designs a memory module that encodes both the input data and the latest memory to produce node representations. This memory module leverages these representations to compute messages and update the node memory accordingly. Mecoin~\cite{li2024efficient} adopts Memory Augmentation for Representation Learning (MRaM), which caches class probability distributions to decouple prototype learning from representation learning, thereby preserving knowledge of seen categories during continual updates. These memory-based design inspire our memory-based module.

\subsection{Contrastive Learning}

% Unlike supervised learning, which relies on manually labeled data, contrastive learning trains models by distinguishing between similar and dissimilar samples, maximizing the similarity of positive pairs while minimizing that of negative pairs. 
% Contrastive learning often employs data augmentation techniques, such as flipping, rotation, and adding noise, to create positive samples. These augmentations help the model focus on core features while ignoring external noise. This learning paradigm is widely used in image processing, as well as text and speech tasks, providing efficient solutions for unsupervised or semi-supervised learning. By significantly improving representation quality, contrastive learning enables better performance on downstream tasks across computer vision, natural language processing, and speech recognition.

% The similarity between all tokens in a sequence and all tokens in other sequences is calculated. For example, for each token in a sequence of length \(L_1\) and batch \(B_1\), the similarity is computed with each token in a sequence of length \(L_2\) and batch \(B_2\). The maximum value is selected as the similarity between each token and the target sequence. 

\begin{figure*}[t]
  \centering
  \includegraphics[width=\textwidth]{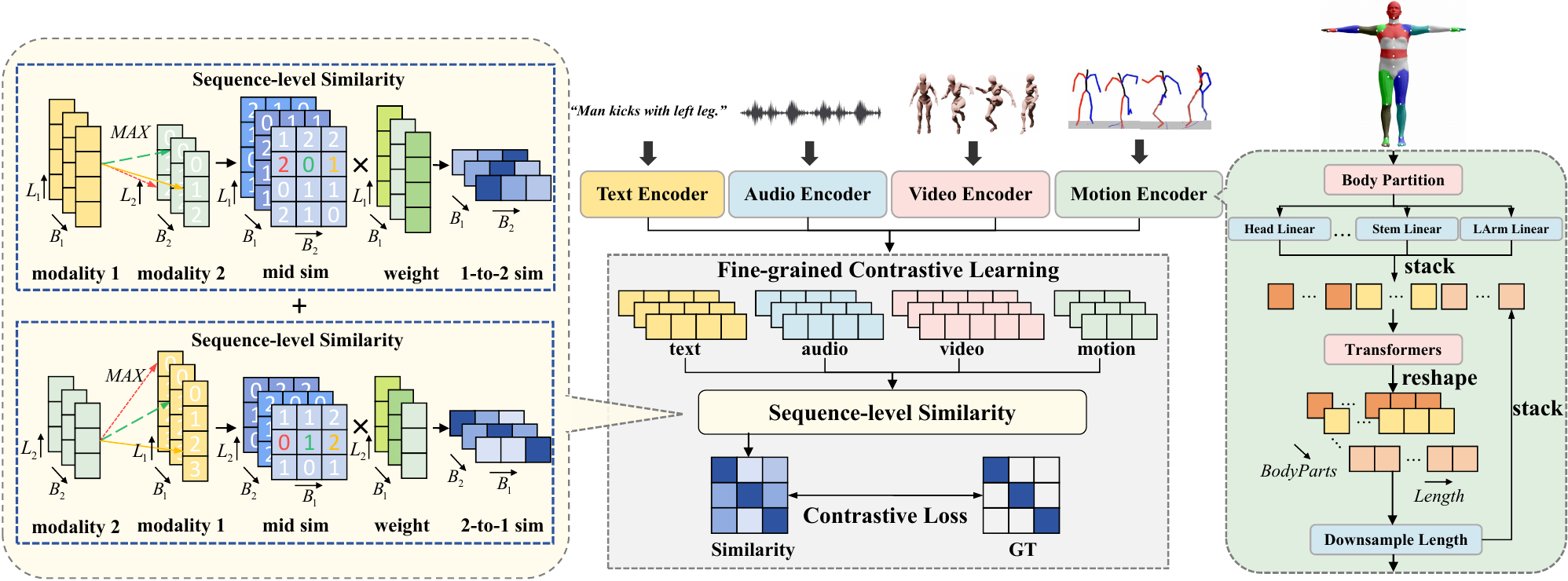}
  \caption{
    \textbf{Pipeline of Our Work.} 
    The text, audio, and video inputs are encoded using pre-trained models to extract feature tokens. Meanwhile, the motion encoder segments human motion by body parts and processes them through a transformer, effectively integrating both body part and temporal information. This design ensures that the alignment process captures both spatial and temporal dependencies. 
    Once inputs from different modalities are encoded, a fine-grained contrastive loss is applied to align them within a joint embedding space. Specifically, for a given token in modality 1 with shape $B_1\times L_1$, we compute its similarity with tokens in modality 2 of shape $B_2\times L_2$ and select the maximum value as its similarity with a sequence in modality 2. This results in a similarity matrix of shape $B_1\times L_1\times B_2$. To account for the varying informativeness of different tokens, we apply a learnable weight, producing a final similarity matrix of dimension $B_1 \times B_2$ between the two modalities. 
    }
    \label{fig:audio_performance}
\end{figure*}

Contrastive learning is a self-supervised learning method aimed at learning meaningful representations from unlabeled data. This approach allows models to capture relationships between data points more effectively. CLIP~\cite{radford2021learning} is a well-known example that leverages contrastive learning to establish connections between images and text. With its openness and strong transferability, CLIP handles diverse tasks without relying on predefined fixed classification labels, making it a representative model in the field of multi-modal learning. Additionally, ImageBind~\cite{girdhar2023imagebind} takes an innovative approach to multi-modal alignment by utilizing image-paired data, enabling the creation of a shared representation space for audio, images, and videos. Similarly, GLIP~\cite{li2022grounded} extends CLIP to open-vocabulary object detection by aligning images with textual descriptions at the region level, enabling the model to detect novel objects without requiring task-specific retraining. Likewise, ActionCLIP~\cite{wang2021actionclip} adapts CLIP for action recognition by integrating temporal information from videos, allowing it to learn robust video-text representations and achieve strong performance in zero-shot and transfer learning scenarios. Moreover, MP-FGVC~\cite{jiang2024mpfgvc} leverages the cross-modal capabilities of the pre-trained CLIP model to improve performance on fine-grained visual classification (FGVC) tasks, representing a pioneering effort to extend the applicability of large-scale vision-language models to FGVC. These advancements demonstrate the versatility of CLIP-based models in various multi-modal learning tasks, highlighting their ability to generalize across different domains through contrastive learning.

% However, when processing sequences of text, audio, and video, ImageBind focuses solely on global information while neglecting the fine-grained information contained within individual units of the sequences. These fine-grained features are crucial for tasks involving multi-modality alignment. To address this limitation, our method introduces an innovative strategy for sequence-level alignment, which will be detailed in the following sections.
% CLIP is a prime example of a model that leverages contrastive learning to establish deep connections between images and text. CLIP's training involves three stages: (1) Contrastive pretraining: learning image-text matching relationships through contrastive learning with large-scale image-text pairs; (2) Create dataset classifier from label text: extracting features from text labels; (3) Use for zero-shot prediction: making predictions on unseen data without additional labeled samples. With its openness and strong transferability, CLIP handles diverse tasks without relying on predefined fixed classification labels, making it a representative model in the field of multimodal learning. 

\section{Method}
% Our main objective is to construct a joint embedding space that aligns motion with the other three modalities: text, video, and audio. Based on this joint space, we can conduct cross-modality retrieval between text and motion, video and motion, as well as audio and motion. At the same time, to achieve fine-grained alignment between motion and other modalities, we have designed a motion encoder with attention to both the temporal and bodily aspects, and a fine-grained contrastive loss that aligns the temporal body part features with other sequential modality features. 
Our primary objective is to construct a joint embedding space that aligns motion with three other modalities: text, video, and audio. This unified space enables cross-modal retrieval between text and motion, video and motion, and audio and motion. To achieve fine-grained alignment, we design a motion encoder that captures both temporal and bodily relationship, along with a fine-grained contrastive loss that aligns temporal body-part features with corresponding sequential features from other modalities.

% The overall model architecture is detailed in Section $A$, and the fine-grained alignment method is introduced in Section $B$.

\subsection{Multi-Modal Data Encoder}
\subsubsection{Motion Encoder}

To effectively capture the spatial and temporal features of human motion, we design a motion encoder that processes body parts individually. This approach is based on the assumption that humans recognize motion by analyzing the relative movements of different body parts. We adopt the motion representation proposed by Guo et al.~\cite{guo2022generating} as the input $M$, which is modeled as a sequence of human poses $\{P_1, P_2, \dots, P_i, \dots\}$. To better encode spatial information, each human pose $P_i$ is decomposed into eight body parts, including the limbs and torso, represented as $\{ b_1, b_2, \dots, b_8 \}$. Each body part $b_k$ consists of a set of two to four joints.

The pose encoder first takes motion $M$ as input and projects the data of each body part into its independent latent space using a set of linear layers. This process enables the model to learn distinct representations for each body part, facilitating more effective motion recognition. Once the features for each body part are embedded, they are stacked together to form a unified feature tensor of shape $\mathbf{e}_m' \in \mathbb{R}^{B \times K \times L_{m}' \times C}$, where $B$ is the batch size, $K$ is the number of body parts, $L_{m}'$ is the length of the input motion feature, and $C$ is the feature dimension. 

The unified feature tensor then undergoes multiple stages of encoding. In each stage, the encoder first applies temporal encoding to capture the sequential nature of the motion data. Next, the temporal and body part dimensions of the feature tensor are swapped, and spatial encoding is applied to model the relationships between different body parts. This is followed by a pooling layer that reduces the temporal sequence length, allowing the model to capture higher-level motion representations over time. This multi-stage processing enables the model to effectively learn both local and global dependencies within the motion sequence, ultimately enhancing its ability to recognize complex motions. In the final stage, the body part and motion length dimensions are reshaped together, producing the output $\mathbf{e}_m \in \mathbb{R}^{B \times KL_m \times C}$, where $L_m$ represents the temporal length of motion feature tokens.

% The feature that were previously mixed are restored to its original dimension-independent state at the end of each encoding stage. 

% To manage variable sequence lengths, a mask is generated to identify valid timesteps, ensuring that the model only attends to meaningful input during training. The encoder then processes the feature tensor with the mask applied, enabling the model to focus on the relevant parts of the sequence while ignoring padded sections. 

% The sentence includes three types of words: the target role (\eg, “a person”), specific motion performed by the role (\eg, “passing a ball”, “runs backwards”), and the attributes that characterize these motions (\eg, “in a counterclockwise motion”). It is important to note that a text description may contain multiple actions.

\subsubsection{Text Encoder}
% Both text and video encoders are built based on the encoders provided by LAVIMO, which also serves as our selected baseline. Both encoders use a Transformer architecture. 
The text input is a sequence of words describing the motion, represented as $ \{ w_1, w_2, \dots, w_{L_t} \} $. These sentences typically contain three key components: the target role (e.g., “a person”), the specific motion performed by the role (e.g., “passing a ball” or “running backward”), and attributes that further characterize the motion (e.g., “in a counterclockwise direction”). Notably, a single text description may encompass multiple actions.

The sequence is first processed by DistilBERT~\cite{sanh2019distilbert} to generate textual tokens. Subsequently, the sequence of word tokens is fed into multiple layers of transformer encoder. The resulting output is denoted as $\mathbf{e}_t \in \mathbb{R}^{B \times L_t \times C}$, where $L_t$ is the number of words.
\begin{equation}
    \label{deqn_ex1a}
    \mathbf{e}_t = \text{Transformer}\big(\text{DistilBERT}(\{ w_1, w_2, \dots, w_{L_t} \} )\big).
\end{equation}
Instead of using a global representation based on the \texttt{<cls>} token, we maintain the token length and adopt a sequence-level alignment approach, ensuring that the full set of text features generated by the transformer encoder is utilized for more precise alignment.

% Then the sequence of text tokens is fed into multiple layers of transformer encoder. We employ a sequence-level alignment approach, opting not to use the global representation based on the \texttt{<cls>} token. Instead, we use the complete text features output by the transformer encoder layers, denoted as $\mathbf{e}_t \in \mathbb{R}^{B \times L_t \times C}$, where $B$ is the batch size, $L_t$ is the number of words, and $C$ is the feature dimension for each group.

\subsubsection{Video Encoder}
We uniformly sample $L_v$ frames from an RGB video, forming the input sequence $\{ I_1, I_2, \dots, I_{L_v} \}$. The video encoder comprises CLIP’s~\cite{radford2021learning} image encoder followed by a temporal transformer encoder. First, the image encoder extracts features from the selected frames. Given the sequential nature of video data, these extracted features are then processed by the temporal transformer encoder, producing $\mathbf{e}_v \in \mathbb{R}^{B \times L_v \times C}$. 
\begin{equation}
    \label{deqn_ex1a}
    \mathbf{e}_v = \text{Transformer}\big(\text{CLIP}(I_1, I_2, \dots, I_{L_v})\big),
\end{equation}
Similar to our approach for text encoding, we avoid using a global token representation and retain the full set of features.

\subsubsection{Audio Encoder}
% To effectively retrieve human motion based on audio instructions, we designed a speech vectorization module that transforms raw audio signals into suitable feature representations. By standardizing feature dimensions across audio samples, this module ensures compatibility with downstream tasks, even with varying sequence lengths.

To enable effective human motion retrieval from audio instructions, we design a speech vectorization module that converts raw audio signals into structured feature representations. This module standardizes feature dimensions across different audio samples, ensuring consistency and compatibility with downstream tasks, even when sequence lengths vary significantly.

Our audio vectorization module is built on the pre-trained WavLM~\cite{chen2022wavlm} model, leveraging the output from its final layer as the vectorized audio representation. Then, the audio signals are represented as a tensor of dimensions $\mathbb{R}^{B\times n_{\text{samples}}}$, where $B$ is the batch size and $n_{\text{samples}}$ is the number of audio samples per batch. To ensure consistency in batch processing, we standardize audio lengths by padding shorter samples with zeros and truncating longer ones to a fixed duration. 

% In the subsequent section, we provide a detailed overview of the audio description generation process. The processed audio descriptions are first converted into audio tokens using WavLM and then mapped to the shared latent space of other modalities through a linear transformation.

\begin{figure}[t]
  \centering
  \includegraphics[width=\columnwidth]{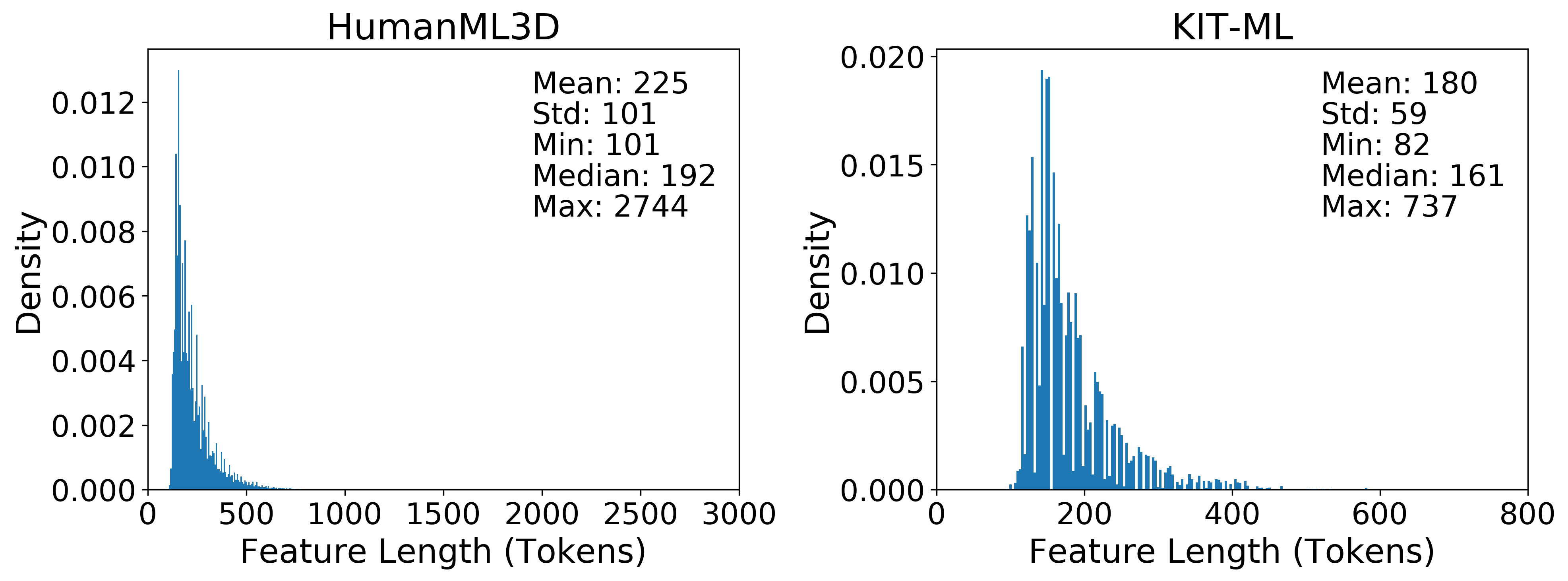}
  \caption{
    \textbf{Distribution of Audio Feature Lengths.} Audio features are extracted from the augmented Oral Datasets, derived from HumanML3D~\cite{guo2022generating} and KIT-ML~\cite{plappert2016kit}, using WavLM~\cite{chen2022wavlm}. The statistical analysis of these feature lengths highlights substantial variability, with some features being exceptionally long. Such variability poses challenges in processing conditional signals, making it more difficult to seamlessly incorporate audio data into later stages of the framework.
    }
    \label{fig:wavlm_feature_length}
\end{figure}

However, we observe that $n_{\text{samples}}$ varies significantly across the dataset, exhibiting a long-tailed distribution. This variability poses challenges in effectively handling audio features for motion retrieval. Specifically, the distribution of audio feature lengths across samples in the KIT-ML~\cite{plappert2016kit} and HumanML3D~\cite{guo2022generating} datasets is shown in Fig.~\ref{fig:wavlm_feature_length}. Directly feeding these lengthy features into the model can lead to high computational complexity and hinder model convergence due to the excessive input size.

A common approach to this issue is appending a special token to each sequence of audio features and processing the entire sequence through a transformer encoder, which aggregates information via self-attention. This token captures global information, allowing the model to handle varying sequence lengths. To reduce computational burden from long raw audio inputs, a CNN can be used for downsampling, condensing the sparse representation into a more compact and semantically enriched form. However, our experiments showed suboptimal performance, likely due to the bottleneck of relying on a single token to represent rich audio features. Additionally, irrelevant information, such as speaker identity, introduces noise, making it harder to extract motion-related semantics.

Instead, we design a memory-retrieval-based module to address this issue. Specifically, we introduce a set of learnable tokens stored in memory, which serve as key-value pairs in the attention mechanism. The key and value are denoted as $K_m$ and $V_m$, respectively. For each input $z$, a fully connected network generates a query vector $Q_z$, which is then used in the attention mechanism to compute a weighted sum of the learnable tokens, producing a refined feature representation. To enhance contextual understanding, positional encoding is added before passing the feature to the next stage of the encoder, allowing for more precise and semantically relevant feature extraction. This process is formulated as follows:
\begin{equation}
    \label{deqn_ex1a}
    \text{Attention}(Q_z, K_m, V_m) = \text{softmax}\left( \frac{Q_z K_m^{\top}}{\sqrt{C}} \right) V_m.
\end{equation}

% This approach not only alleviates the computational burden but also ensures the extraction of rich motion-related semantic information, enabling more efficient processing in the downstream stages. It compresses the temporal dimension and enhances the stability of the representations. The resulting audio features $\mathbf{e}_a$ have dimensions of $B \times L_a \times C$, where $L_a$ represents the length of the compressed audio features. The steps involved in this module are illustrated in Fig.~\ref{fig:audio_processing}.

\begin{figure}[t]
  \centering
  \includegraphics[width=\columnwidth]{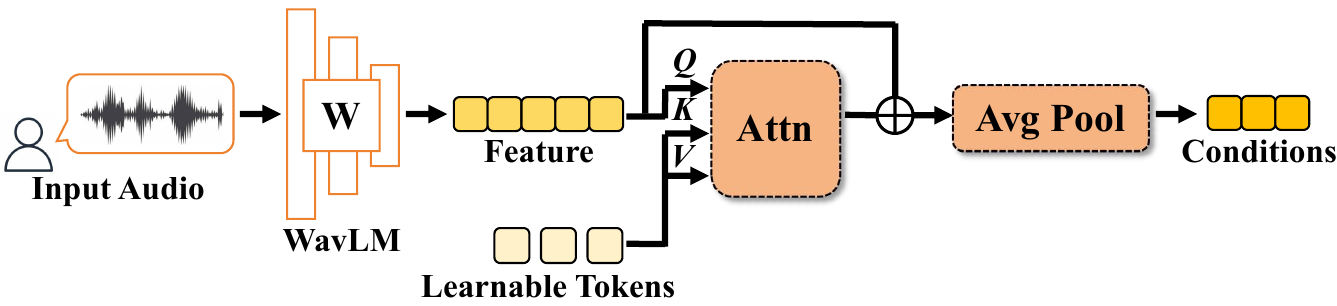}
  \caption{
    \textbf{Overview of the Audio Processing.} WavLM~\cite{chen2022wavlm} extracts audio features, which are then processed through a memory-retrieval-based module designed to standardize input audio conditions of significantly varying lengths. By converting all audio signals into a uniform length, this module ensures seamless integration with the subsequent stages of the pipeline.
    }
    \label{fig:audio_processing}
\end{figure}

This approach reduces computational burden while preserving rich motion-related semantic information, facilitating more efficient downstream processing. By compressing the temporal dimension, it enhances the stability of the representations. The resulting audio features, denoted as $\mathbf{e}_a\in \mathbb{R}^{B \times L_a \times C}$, where $L_a$ represents the length of the compressed audio features. The steps involved in this module are illustrated in Fig.~\ref{fig:audio_processing}.

\subsection{Multi-Modal Sequence-level Alignment}
We utilize contrastive learning to construct a joint embedding space that aligns the four modalities. This alignment is achieved by minimizing the following loss function:
\begin{equation}
\label{deqn_ex1a}
L_{\text{align}} = L^{mt}_{\text{align}} + L^{mv}_{\text{align}} + L^{ma}_{\text{align}} + L^{tv}_{\text{align}} + L^{ta}_{\text{align}} + L^{va}_{\text{align}},
\end{equation}
where $L^{xy}_{\text{align}}$ denotes the alignment loss between modality $x$ and modality $y$, ensuring that their features are well-aligned in the shared embedding space.

% Taking text and motion as an example, a text description paired with a matching motion is defined as a positive sample pair, while a text description paired with a non-matching motion is considered a negative sample pair. We utilize a bidirectional Kullback–Leibler (KL) divergence loss function to bring positive sample pairs closer and push negative sample pairs apart:
In the case of text and motion alignment, a text description paired with a matching motion is treated as a positive sample pair, while a text description paired with a non-matching motion is considered a negative sample pair. To enforce alignment, we employ a bidirectional Kullback–Leibler (KL) divergence loss function, which pulls positive sample pairs closer while pushing negative sample pairs apart:
\begin{equation}
    \label{deqn_ex1a}
    L_{\text{align}}^{xy} = KL\left(\mathbf{S}_{\text{pred}}^{x2y}, \mathbf{S}_{\text{target}}\right) + KL\left(\mathbf{S}_{\text{pred}}^{y2x}, \mathbf{S}_{\text{target}}^{\top}\right),
\end{equation}
where $\mathbf{S}_{\text{pred}}^{x2y}$ and $\mathbf{S}_{\text{pred}}^{y2x}$ denote the similarity matrices from modality $x$ to modality $y$ and that from modality $y$ to modality $x$, respectively. These matrices are computed as follows:
\begin{align}
    \label{deqn_ex1a}
    \mathbf{S}_{\text{pred}}^{x2y}(i, j) &= \frac{\exp(h(\mathbf{e}^i_x, \mathbf{e}^j_y) / \tau)}{\sum_{k=1}^B \exp(h(\mathbf{e}^i_x, \mathbf{e}^k_y) / \tau)}, \\
    \mathbf{S}_{\text{pred}}^{y2x}(i, j) &= \frac{\exp(h(\mathbf{e}^i_y, \mathbf{e}^j_x) / \tau)}{\sum_{k=1}^B \exp(h(\mathbf{e}^i_y, \mathbf{e}^k_x) / \tau)},
\end{align}
where $\tau$ is the learnable temperature parameter, the indices $(i, j)$ denotes the indices of a specific element within the respective matrix, and $h(\mathbf{e}^i_x, \mathbf{e}^j_y)$ represents the similarity between cross-modal sequences.

% Each score is determined by first maximizing $\mathbf{S}_{\text{pred}}^{x2y}$ along one matrix dimension and then averaging along the other dimension.

% For similarity calculation, instead of using global representations of the aligned text and motion, we compute fine-grained similarity between tokens from one modality and tokens from the other modality. The fine-grained similarity matrix is calculated step by step, and the overall similarity score is obtained by summing the bidirectional scores. 

% Instead of relying on global alignment, this mechanism
% selects the maximum similarity score for each token by
% comparing it with all the tokens in the corresponding sequence,
% effectively capturing critical details across modalities. 

Previous works compute similarity between two modalities using a single global token feature, which can lead to information loss in the global representation. To address this, we introduce a fine-grained similarity function that captures relationships between feature token sequences, defined as:
\begin{equation}
    \label{deqn_ex1a} %f^i_{x,\theta}(e_x)
    h(\mathbf{e}_x, \mathbf{e}_y) = \frac{1}{2} \sum_{i=1}^{L_x} \text{w}^i_x \max_{j=1}^{L_y} \big<{\text{e}_x^i}, \text{e}_y^j\big> + \frac{1}{2} \sum_{j=1}^{L_y} \text{w}^j_y \max_{i=1}^{L_x} \big<{\text{e}_y^j}, \text{e}_x^i\big>,
\end{equation}
where $<\cdot, \cdot>$ denotes the dot product between two tokens, and $\mathbf{w}_x$ and $\mathbf{w}_y$ represent the weight vectors for modalities $x$ and $y$, respectively. These weights are obtained by passing each modality feature through independent linear layers and normalizing them using the softmax function. The ablation study in Tab.~\ref{tab:sim_ablation} compares the unbiased mean similarity selection with the proposed maximum similarity selection. The results show that maximum similarity consistently yields better retrieval performance, indicating that focusing on the most aligned token pair more effectively captures the dominant semantic correspondence between multi-modal pairs.

% Given that different tokens are not equally informative, we use learnable weights instead of equal weighting. These weights are obtained by feeding each modality feature into independent linear layers, and normalized by softmax function. This process can be formulated as:
% \begin{equation}
% \label{deqn_ex1a}
% \begin{split}
% \mathbf{w}_{x} = f_{x,\theta}(\mathbf{e}_x)
% \end{split}
% \end{equation}
% The inputs to the similarity function are two components, one from $\mathbf{e}_x$ with dimensions \( L_x \times C \) and the other from $\mathbf{e}_y$ with dimensions \( L_y \times C \). 

The target similarity matrix $S_{\text{target}}$ is formulated as:
\begin{align}
    \label{deqn_ex1a}
    \mathbf{S}_{\text{target}}(i, j) =
    \begin{cases} 
    1, & \text{if } i=j, \\
    0, & \text{otherwise}.
\end{cases}
\end{align}

% Other modalities often contain rich contextual information that can supplement details not easily captured in the motion modality. For instance, text may include descriptions of speed, such as “slowly”. To leverage this information, we concatenate the four modality features output by their encoders and process them using a transformer encoder. The processed motion features are then passed to a reconstruction transformer decoder for rebuilding. The above process can be formulated as:
% Other modalities often contain rich contextual information that can complement details missing from the motion modality. For example, text may describe speed with words like “slowly.” To utilize this information, we first randomly mask motion tokens and obtain $\mathbf{\hat e}_m$. It will be concatenated with the other three modalities' token features and pass through a transformer encoder to recover the masked motion tokens. This process can be formulated as:

Other modalities often provide rich contextual information that complements details missing from the motion modality. For example, text may describe speed using words like “slowly.” To leverage this information, we first randomly mask motion tokens to obtain $\mathbf{\hat e}_m$. The masked motion tokens are then concatenated with token features from the other three modalities and passed through a transformer encoder to reconstruct the missing motion information. This process can be formulated as:
\begin{equation}
    \label{deqn_ex1a}
    \tilde{\mathbf{e}}_m = \text{Transformer}([\mathbf{e}_t , \mathbf{e}_v , \mathbf{e}_a , \mathbf{\hat e}_m]).
\end{equation}
Using the recovered motion tokens, we define the motion reconstruction loss as:
\begin{equation}
    L_{\text{recon}} = \|\tilde{\mathbf{e}}_m - \mathbf{e}_m\|_2.
\end{equation}

The final training loss for our model can be summarized as follows:
\begin{equation}
\label{deqn_ex1a}
L = L_{\text{align}} + \lambda_{\text{recon}} \cdot L_{\text{recon}},
\end{equation}
where the $\lambda_{\text{recon}}$ is the weight of reconstruction loss.

\begin{figure}[t]
  \centering
  \includegraphics[width=0.5\textwidth]{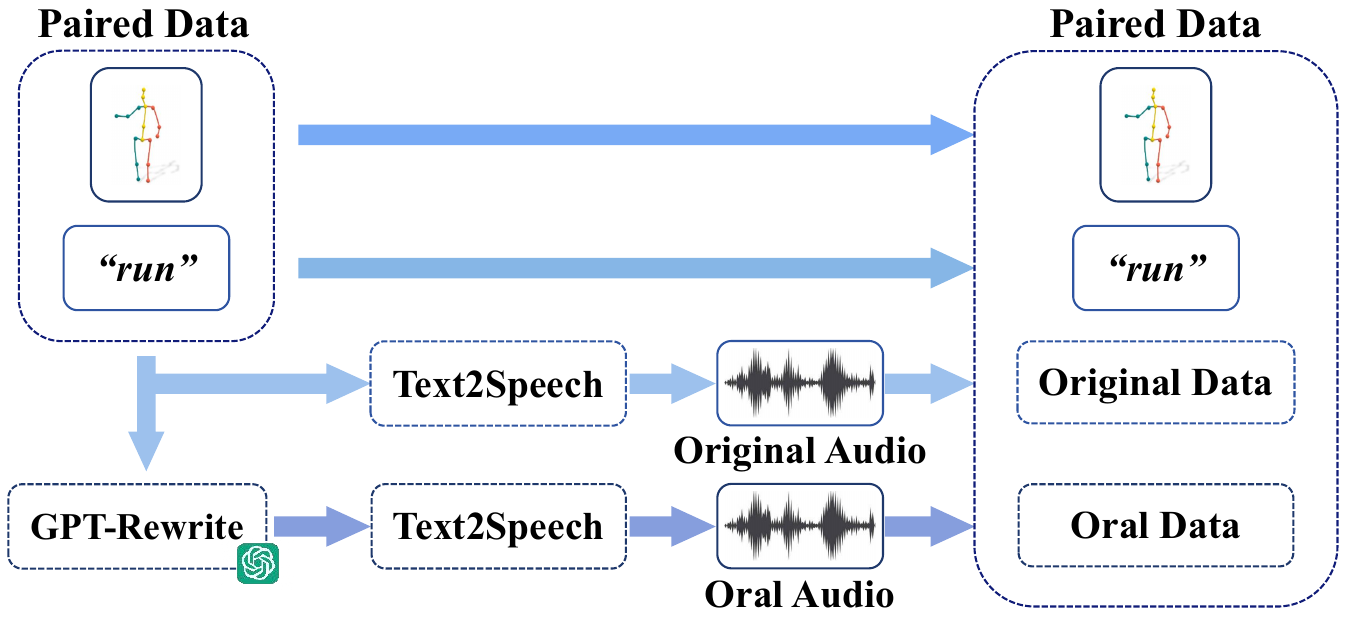}
  \caption{
    \textbf{Dataset Augmentation with Audio Modality.} 
    The text data from the KIT-ML~\cite{plappert2016kit} and HumanML3D~\cite{guo2022generating} datasets are processed using the text-to-speech model Tortoise~\cite{betker2023better} to generate audio signals with randomly assigned speaker identities, forming \textbf{Original Dataset}. Additionally, we use ChatGPT-3.5~\cite{achiam2023gpt} to rewrite these texts into a more conversational, spoken style. The rewritten texts are then converted into audio signals, resulting in \textbf{Oral Dataset}.
    }
    \label{fig:audio_generation}
\end{figure}

\subsection{Motion Dataset Augmentation with Audio Modality}
\label{sec:data-aug}

There is currently no dedicated audio-motion dataset for retrieving human motion based on audio instructions. To address this, we selected a suitable text-to-speech model and synthesized audio from existing text-motion datasets, creating an audio-motion paired dataset to support future research on audio-conditioned motion studies. Among the tested models, Tortoise~\cite{betker2023better} performed best, producing speech that closely aligns with text descriptions while preserving natural vocal traits such as timbre, pauses, and emphasis. Its probabilistic generation framework enables the simulation of diverse speaker voices, enhancing output variety and adaptability across different scenarios. We utilize this capability to generate audio from random speakers, increasing the dataset's diversity.

The two datasets we employ—HumanML3D~\cite{guo2022generating} and KIT-ML~\cite{plappert2016kit}—provide paired textual descriptions for each motion sequence. We employ Tortoise to synthesize corresponding audio clips from the provided descriptions, generating diverse vocal characteristics while maintaining accurate and consistent alignment with the corresponding motion data. Additionally, we employ ChatGPT-3.5~\cite{achiam2023gpt} to paraphrase the original descriptions into more conversational expressions, which are then used to generate corresponding audio. This augmentation enriches both the textual and auditory diversity while maintaining semantic consistency. Examples are provided in Tab.~\ref{tab:gpt_rewritten}.

% In order to make the dataset more applicable to real-world scenarios, we seeks to adjust the descriptive texts to adopt a more conversational style, thus narrowing the gap between the written descriptions and the type of spoken language input that may be encountered. Among the available large language models, ChatGPT has shown outstanding performance in understanding English natural language and following instructions. Given its outstanding performance in understanding English natural language and following instructions, we employ ChatGPT 3.5~\cite{achiam2023gpt} to rephrase the motion descriptions. Tab.~\ref{tab:gpt_rewritten} presents several examples of rewritten motion descriptions.

\begin{table}[h]
    \centering
    % \vspace{-10pt}
    \caption{
    % The original text descriptions from the \textbf{Original Datasets} were rewritten into a more conversational style, forming the augmented \textbf{Oral Datasets}. This transformation aims to better align the dataset with natural spoken language, enhancing its applicability to real-world voice-based interaction scenarios. 
    The original text descriptions from the \textbf{Original Dataset} were rewritten into a more conversational style, creating the augmented \textbf{Oral Dataset}. This adjustment improves alignment with natural spoken language, enhancing its suitability for real-world voice-based interactions.
    }
    \label{tab:gpt_rewritten}
    \begin{tabularx}{\columnwidth}{|X|X|}
        \hline
        \textbf{Original Datasets} & \textbf{Oral Datasets} \\ 
        \hline
        Someone is folding their arms.
        & (1) Fold your arms. \\ 
        & (2) Cross your arms on your chest. \\ 
        & (3) Place your hands on your shoulders, then by your side. \\ 
        \hline
        A person shifts from foot to foot, glancing around his surroundings. 
        & (1) Shift your weight from one foot to the other while scanning your surroundings. \\
        & (2) Rock back and forth on your feet, glancing around. \\ 
        & (3) Move side to side on your feet, checking out your surroundings. \\ 
        \hline
        A person throws something with their right hand then kicks with their right foot before catching the object with both hands. 
        & (1) Throw something with your right hand, kick with your right foot, then catch the object with both hands. \\ 
        & (2) Can you throw with your right hand, kick with your right foot, and catch with both hands? \\ 
        & (3) Please throw using your right hand, kick with your right foot, and catch the object. \\ 
        \hline
    \end{tabularx}
    % \vspace{2pt}
\end{table}

We specifically designed prompts to ensure that the spoken text generated by the model retains the core meaning of the original content. Additionally, the model generates a wider variety of sentence structures and vocabulary for spoken text, reflecting the characteristics of conversational language. Using these revised descriptions, the Tortoise model is subsequently employed to generate the corresponding audio. As a result, this procedure produces 12,696 oral audio-motion pairs for the KIT-ML~\cite{plappert2016kit} dataset and 87,384 pairs for HumanML3D~\cite{guo2022generating}. To support the next audio processing models, we saved all synthesized audio instructions in mono WAV format with sampling rate of 16 kHz, and the synthesis process is accelerated using GPUs. The steps are illustrated in Fig.~\ref{fig:audio_generation}.

% \subsection{Contrastive Learning}

\section{Experiment}
% We conduct experiments on the HumanML3D~\cite{guo2022generating} and KIT-ML~\cite{plappert2016kit} datasets. The results show that in both text-to-motion and video-to-motion retrieval tasks, our model outperforms previous methods. Additionally, we test our model on both the original and oral audio datasets to validate the necessity of constructing the oral audio dataset.
% \subsubsection{\textbf{Datasets}}
% In our study, we use the HumanML3D~\cite{guo2022generating} and KIT-ML~\cite{plappert2016kit} datasets to conduct our experiments. The HumanML3D dataset, which contains 14,616 motions and 44,970 corresponding textual annotations, is the largest 3D human motion dataset with annotated text. The motion data, which originates from AMASS~\cite{mahmood2019amass} and HumanAct12~\cite{guo2020action2motion}, undergoes preprocessing that includes downsampling to 20 FPS, trimming sequences longer than 10 seconds, and aligning them to the Z+ axis. Every motion is paired with at least three descriptions. Furthermore, the text annotations consist of 5,371 unique words. 80\% of the dataset is for training, 5\% is for validation, and 15\% is for testing. As for the KIT-ML dataset, it contains 3,911 motion sequences and 6,278 textual annotations, with a total of 1,623 unique words. Each motion is paired with one to four descriptions and is downsampled to 12.5 FPS. Additionally, 80\% of the dataset is used for training, 5\% for validation, and 15\% for testing.
\textbf{Datasets.} We evaluate our approach on HumanML3D~\cite{guo2022generating} and KIT-ML~\cite{plappert2016kit} datasets. HumanML3D comprises 14,616 motion sequences and 44,970 corresponding textual annotations. The motion data originates from AMASS~\cite{mahmood2019amass} and HumanAct12~\cite{guo2020action2motion} and undergoes preprocessing, including downsampling to 20 FPS, trimming sequences longer than 10 seconds, and aligning them to the Z+ axis. Each motion is paired with at least three descriptions, and the text annotations contain a total of 5,371 unique words. Similarly, the KIT-ML dataset consists of 3,911 motion sequences and 6,278 textual annotations, with 1,623 unique words. Each motion is paired with one to four descriptions and is downsampled to 12.5 FPS. We follow the same train/test data split as prior work~\cite{guo2022generating,plappert2016kit}.

\textbf{Implementation Details.} 
% All experiments are carried out using 8 NVIDIA A6000 GPUs with PyTorch~\cite{paszke2017automatic}. The motion encoder consists of a 4-layer transformer encoder, with an average pooling layer inserted between every two transformer layers to reduce the channel dimensions. The text encoder is built upon DistilBERT~\cite{sanh2019distilbert} and is enhanced with a temporal transformer to account for word embedding positions. DistilBERT is fine-tuned during the training process. The video encoder utilizes a ViT-B/32 with 12-layers, complemented by a temporal transformer with 6 layers to capture temporal information. In accordance with~\cite{wu2023revisiting}, we initialize DistilBERT and the CLIP image encoder, both pre-trained on the Kinetics-400~\cite{carreira2018short} dataset. We configure $L_v$ to 8, set the latent dimension $C$ to 512, and assign 0.1 to $\lambda_{\text{recon}}$. 
% The motion encoder consists of a four-layer transformer encoder, with an average pooling layer inserted between every two transformer layers to reduce temporal size. The text encoder is based on DistilBERT~\cite{sanh2019distilbert}, enhanced with a temporal transformer to incorporate sequential dependencies. The video encoder employs a ViT-B/32~\cite{dosovitskiy2020image} with 12 layers, supplemented by a six-layer temporal transformer to capture temporal dependencies. Following~\cite{yin2024tri}, we initialize with the image encoder using pre-trained weights from CLIP~\cite{radford2021learning}. 
The motion encoder consists of a four-layer transformer encoder, with an average pooling layer inserted between every two layers to reduce the temporal size. The text encoder is built upon DistilBERT~\cite{sanh2019distilbert}, enhanced with a two-layer transformer to capture sequential dependencies. The video encoder utilizes a ViT-B/32~\cite{dosovitskiy2020image} as the image encoder, further augmented by a six-layer transformer to model temporal relationships. Following~\cite{yin2024tri}, we initialize the image encoder with pre-trained weights from CLIP~\cite{radford2021learning}.

% During training, we set the batch size to 64 and training epochs to 200. All experiments are conducted using eight NVIDIA A6000 GPUs. We set the latent dimension $C = 512$, and $\lambda_{\text{recon}} = 0.1$. We employ the AdamW optimizer~\cite{oord2018representation} with an initial learning rate of 1e-4, which linearly decays to 1e-5 after the first 100 epochs. During data augmentation, an image is first resized at a random position, and a $256\times256$ pixel crop is extracted. Subsequently, the cropped image undergoes various transformations, including random color jittering, random conversion to grayscale, application of Gaussian blur, and random horizontal flipping implemented through RandAugment~\cite{cubuk2020randaugment}. Our 2-modality, 3-modality, and 4-modality versions share the same configurations, differing only in the contrastive learning and modality fusion involving varying types of modalities.

All experiments are conducted using eight NVIDIA A6000 GPUs. We train the model for 200 epochs with a batch size of 64. The latent dimension is set to $C = 512$, and the reconstruction loss weight is $\lambda_{\text{recon}} = 0.1$. All alignment losses are assigned an equal weight of 1, as L2 normalization applied to embeddings ensures they are on a similar scale, resulting in stable and balanced optimization across modalities. We use the AdamW optimizer~\cite{oord2018representation} with an initial learning rate of 1e-4, which linearly decays to 1e-5 after the first 100 epochs. For data augmentation, images are first randomly resized and cropped to $256\times256$ pixels. The cropped images then undergo a series of transformations, including random color jittering, grayscale conversion, Gaussian blur, and horizontal flipping, implemented via RandAugment~\cite{cubuk2020randaugment}. Our 2-modality, 3-modality, and 4-modality versions share the same overall configurations, differing only in contrastive learning strategies and modality fusion mechanisms.

\textbf{Evaluation Metrics.}
We evaluate retrieval performance with standard metrics, including recall at different ranks (e.g., R@1, R@3) across text-motion, video-motion, and audio-motion tasks. The higher the R-precision, the greater the retrieval accuracy. In addition, we use median rank (MedR) to evaluate retrieval accuracy, where MedR represents the median rank of the ground-truth results, with a lower rank indicating higher retrieval precision. We further consider three evaluation protocols: (i) \textbf{All}, which evaluates retrieval performance over the entire test set; (ii) \textbf{All with threshold}, where a retrieval is considered correct if the similarity between the retrieved motion and the ground-truth exceeds a predefined threshold (set to 0.80 in our experiments); and (iii) \textbf{Small batches}, which averages performance over randomly sampled batches of 32 cross-modal pairs.
% The four evaluation protocols we use are listed below: (i) \textbf{All} refers to using the entire test dataset as the retrieval dataset. (ii) \textbf{All with threshold} sets a threshold: if the similarity between the retrieved motion and the ground-truth motion exceeds the threshold, the result is deemed correct. This protocol is designed to address the accuracy drop caused by motions similar to the ground-truth description being classified as negatives. (iii) \textbf{Dissimilar subset} utilizes a curated subset as the test set. This subset consists of 100 pairs of samples, each of which is clearly different. Compared to the previous protocol, using this protocol makes it relatively easier to retrieve the correct motion. (iv) \textbf{Small batches} randomly selects 32 pairs of motion-text and evaluates the average performance.

\subsection{Comparison to Prior Works}
% We compare our method with prior works, including TEMOS~\cite{petrovich2022temos}, TMR~\cite{petrovich2023tmr} and LAVIMO~\cite{yin2024tri}, on text-to-motion and video-to-motion retrieval tasks, and also present its performance on the audio-to-motion retrieval task.

% For the text-to-motion retrieval task, we present two different versions of the framework: the 3-modality version trained on text, motion, and video modalities, and the 4-modality version trained on all four modalities. On the HumanML3D~\cite{guo2022generating} dataset, our 4-modality framework outperforms the state-of-the-art LAVIMO~\cite{yin2024tri} method in both text-to-motion and motion-to-text retrieval tasks (details are provided in Tables~\ref{tab:h3d}). The text-to-motion retrieval results on the HumanML3D dataset are shown in Fig.~\ref{fig:text_comparison}. On the KIT-ML dataset, we observe the same phenomenon as in the HumanML3D dataset (shown in Table~\ref{tab:kit}). Our four-modality framework also outperforms the three-modality version on both two datasets, demonstrating that the additional audio modality indeed enhances the alignment between text and motion modalities. 

We compare our method with prior works, including TEMOS~\cite{petrovich2022temos}, MotionCLIP~\cite{tevet2022motionclip}, the approach proposed by Guo~\cite{guo2022generating}, TMR~\cite{petrovich2023tmr}, and LAVIMO~\cite{yin2024tri}, on text-motion task. We further compare our method with the state-of-the-art model LAVIMO on video-motion retrieval, while also evaluating our performance on the audio-motion retrieval task.

\begin{figure*}[h!]
    \centering
    \includegraphics[width=\textwidth]{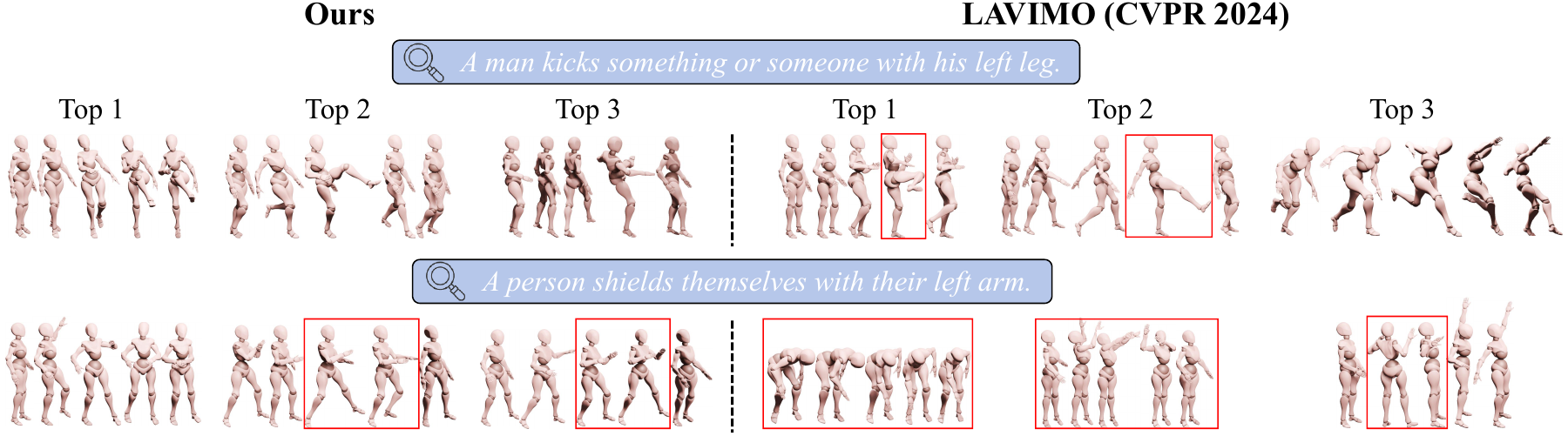}
    \caption{\textbf{Qualitative Comparison of Text-to-Motion Retrieval on the HumanML3D Dataset.} We compare our results with LAVIMO~\cite{yin2024tri}. In the first row, our model accurately retrieves the motion of kicking the right leg, while the comparison model retrieves a motion involving the left leg. A similar pattern is observed in the second row. Overall, our model demonstrates more precise retrieval, with its Top-2 and Top-3 results better aligning with the text description.
    % For text-to-motion retrieval task, we compare our results with LAVIMO~\cite{yin2024tri}. In the first row, our model successfully retrieves the motion of kicking the right leg, while the comparison model retrieves the left leg instead. Overall, our model's top2 and top3 retrieval results are closer to the correct motion compared to the comparison models.
    }
    \label{fig:text_comparison}
\end{figure*}

For text-motion retrieval, we introduce two versions of our framework: a 3-modal version trained on text, motion, and video modalities, and a 4-modal version that additionally incorporates audio. On the HumanML3D~\cite{guo2022generating} dataset (results shown in Tab.~\ref{tab:h3d}), our 4-modal framework outperforms the state-of-the-art LAVIMO~\cite{yin2024tri} under the All protocol in both text-to-motion (43.83 vs. 33.67 in R@10) and motion-to-text (43.74 vs. 36.55 in R@10) retrieval. Our 3-modal framework also surpasses LAVIMO by a significant margin under the All protocol in text-motion retrieval. These results highlight the effectiveness of our multi-modal approach in leveraging complementary information to enhance fine-grained alignment in the embedding space. Fig.~\ref{fig:text_comparison} further illustrates qualitative text-motion retrieval results, showing improved alignment of the Top-2 and Top-3 retrieved motions with the text description. A similar trend is observed on the KIT-ML dataset (results shown in Tab.~\ref{tab:kit}), where our four-modal approach achieves noticeable improvements over LAVIMO in both text-to-motion (60.22 vs. 49.80 in R@10) and motion-to-text (58.04 vs. 53.32 in R@10) retrieval. Additionally, our model outperforms prior state-of-the-art approaches on both HumanML3D and KIT-ML datasets across the All with threshold and Small batches protocols.

% Our 4-modal framework outperforms the state-of-the-art LAVIMO~\cite{yin2024tri} in both text-to-motion and motion-to-text retrieval (detailed results are provided in Table~\ref{tab:h3d}). Fig.~\ref{fig:text_comparison} illustrates the text-to-motion retrieval results on the HumanML3D dataset. A similar trend is observed on the KIT-ML dataset, as shown in Table~\ref{tab:kit}. Additionally, our 4-modal framework consistently outperforms the three-modality version across both datasets, demonstrating that incorporating the audio modality further enhances the alignment between text and motion representations.

\begin{table*}[h]
\centering
\caption{
\textbf{Retrieval Results on HumanML3D~\cite{guo2022generating}.} Our 4-modal version outperforms pervious methods and our 3-modal version, demonstrating the effectiveness of our multi-modal framework with fine-grained alignment. The best results are in \textbf{bold}.}
\resizebox{0.9\textwidth}{!}{%
\begin{tabular}{@{}l|l|ccccc|ccccc@{}}
\toprule
\multirow{2}{*}{Protocol} & \multirow{2}{*}{Methods} & R@1↑ & R@3↑ & R@5↑ & R@10↑ & MedR↓ & R@1↑ & R@3↑ & R@5↑ & R@10↑ & MedR↓ \\
                          &        & \multicolumn{5}{c|}{Text-motion retrieval} & \multicolumn{5}{c}{Motion-text retrieval} \\
\midrule
\multirow{6}{*}{(a) All} & TEMOS~\cite{petrovich2022temos}    & 2.12  & 5.87 & 8.26 & 13.52 & 173.00 & 3.86  & 6.94 & 9.38 & 14.00 & 183.25 \\
                         & Guo et al.~\cite{guo2022generating}   & 1.80 & 4.79 & 7.12 & 12.47 & 81.00 & 2.92  & 6.00 & 8.36 & 12.95 & 81.50 \\
                         & MotionCLIP~\cite{tevet2022motionclip}  & 2.33  & 8.93 & 12.77 & 18.14 & 103.00 & 5.12 & 8.35 & 12.46 & 19.02 & 91.42 \\
                         & TMR~\cite{petrovich2023tmr}   & 5.68  & 14.04 & 20.34 & 30.94 & 28.00 & 9.95  & 17.95 & 23.56 & 32.69 & 28.50 \\
                         & LAVIMO(3-modal)~\cite{yin2024tri}   & 6.37  & 15.60 & 21.95 & 33.67 & 24.00 & 9.72  & 18.73 & 25.00 & 36.55 & 22.50 \\
                         & Ours(3-modal)     & 8.78  & 19.19 & 27.10 & 40.69 & 16.00 & 9.32  & 20.78 & 27.64 & 41.61 & 15.00 \\
                         & Ours(4-modal)     & \textbf{9.41}  & \textbf{21.95} & \textbf{30.07} & \textbf{43.83} & \textbf{14.00} & \textbf{10.03}  & \textbf{22.75} & \textbf{30.40} & \textbf{43.74} & \textbf{14.00} \\
                         \midrule
\multirow{6}{*}{(b) All with threshold} & TEMOS~\cite{petrovich2022temos}  & 5.21 & 11.14 & 15.09 & 22.12 & 79.00  & 5.48 & 9.00 & 12.01 & 17.10 & 129.00 \\
                         & Guo et al.~\cite{guo2022generating}   & 5.30 & 10.75 & 14.59 & 22.51 & 54.00 & 4.95 & 8.93 & 11.64 & 16.94 & 69.50 \\
                         & MotionCLIP~\cite{tevet2022motionclip}  & 7.22 & 13.48 & 19.07 & 23.65 & 55.00  & 7.10 & 13.57 & 20.04 & 25.44 & 53.87 \\
                         & TMR~\cite{petrovich2023tmr}   & 11.60 & 20.50 & 27.72 & 38.52 & 19.00  & 13.20 & 22.03 & 27.65 & 37.63 & 21.50  \\
                         & LAVIMO(3-modal)~\cite{yin2024tri}  & 12.94 & 23.63 & 30.13 & 42.46 & 16.00  & 13.89 & 23.83 & 29.93 & 41.09 & 17.50  \\
                         & Ours(3-modal)             & 14.85 & 33.08 & 43.50 & 56.96 & 7.00  & 16.27 & 33.38 & 41.82 & 53.16 & 9.00  \\
                         & Ours(4-modal)             & \textbf{18.44} & \textbf{37.35} & \textbf{46.51} & \textbf{60.77} & \textbf{6.00} & \textbf{20.03} & \textbf{37.35} & \textbf{45.13} & \textbf{57.21} & \textbf{7.00} \\  \midrule
\multirow{6}{*}{(c) Small batches~\cite{guo2022generating}} & TEMOS~\cite{petrovich2022temos}   & 40.49 & 61.14 & 70.96 & 84.15 & 2.33  & 39.96 & 61.79 & 72.40 & 85.89 & 2.33  \\
                         & Guo et al.~\cite{guo2022generating}   & 52.48 & 80.65 & 89.86 & 96.58 & 1.39  & 52.00 & 81.11 & 89.87 & 96.78 & 1.38  \\
                         & MotionCLIP~\cite{tevet2022motionclip}  & 46.24 & 68.93 & 80.47 & 91.35 & 1.88  & 44.76 & 65.22 & 77.83 & 90.19 & 2.03  \\
                         & TMR~\cite{petrovich2023tmr}   & 67.16 & 86.81 & 91.43 & 95.36 & 1.04  & 67.97 &  86.35 & 91.70 & 95.27 & 1.03  \\
                         & LAVIMO(3-modal)~\cite{yin2024tri}  & 68.58 & 85.02 & 88.77 & 92.58 & \textbf{1.01}  & 68.64 & 85.52 & 88.76 & 92.82 & \textbf{1.01}  \\ 
                         & Ours(3-modal)             & 70.81 & 87.66 & 90.88 & 96.99 & \textbf{1.01}  & 68.80 & 85.36 & 89.04 & 95.06 & \textbf{1.01}  \\
                         & Ours(4-modal)    & \textbf{71.98} & \textbf{88.29} & \textbf{92.30} & \textbf{97.66} & \textbf{1.01}  & \textbf{71.27} & \textbf{86.83} & \textbf{91.22} & \textbf{98.83} & \textbf{1.01} \\
                         \midrule
                         & & \multicolumn{5}{c|}{Video-to-Motion Retrieval} & \multicolumn{5}{c}{Motion-to-Video Retrieval} \\
                         \midrule
\multirow{3}{*}{(a) All} & LAVIMO(3-modal)~\cite{yin2024tri}   & 39.35  & 67.41 & 78.50 & 90.46 & 2.00 & 48.09 & 75.74 & 84.69 & 92.89 & 2.00 \\
                         & Ours(3-modal)    & 60.10 & 84.90 & 91.38 & 95.85 & \textbf{1.00} & 60.85 & 86.07 & 92.09 & 96.31 & \textbf{1.00} \\
                         & Ours(4-modal)     & \textbf{64.78}  & \textbf{88.33} & \textbf{93.56} & \textbf{96.65} & \textbf{1.00} & \textbf{66.54}  & \textbf{88.79} & \textbf{94.48} & \textbf{97.53} & \textbf{1.00} \\
                         \midrule
& & \multicolumn{5}{c|}{Audio-to-Motion Retrieval} & \multicolumn{5}{c}{Motion-to-Audio Retrieval} \\    \midrule
\multirow{1}{*}{(a) All} & Ours(4-modal)  & 16.19 & 31.91 & 40.28 & 53.20 & 9.00 & 17.40 & 33.12 & 41.03 & 51.48 & 10.00 \\ 
\bottomrule
\end{tabular}%
}
\label{tab:h3d}
\end{table*}

% For video-motion retrieval, the evaluation results in Tab.~\ref{tab:h3d} and Tab.~\ref{tab:kit} consistently show that our 4-modal framework significantly outperforms the performance of LAVIMO on both datasets. Furthermore, the 4-modal model also outperforms our 3-modal model, demonstrating that the audio modality aids in the alignment of the other three modalities.
For video-to-motion retrieval, the evaluation results in Tab.~\ref{tab:h3d} and Tab.~\ref{tab:kit} show that our 4-modal framework significantly outperforms LAVIMO on both datasets. Moreover, our 4-modal model also surpasses the 3-modal version, highlighting the contribution of the audio modality in enhancing the alignment in our multi-modal framework.

\begin{table*}[h]
\centering
\caption{
\textbf{Retrieval Results on KIT-ML~\cite{plappert2016kit}.} Our 3-modal model consistently outperforms prior methods under both additional experimental methods. Notably, the 4-modal variant delivers even more substantial improvements across multiple benchmarks. The best results are in \textbf{bold}.
}
\resizebox{0.9\textwidth}{!}{%
\begin{tabular}{@{}l|l|ccccc|ccccc@{}}
\toprule
\multirow{2}{*}{Protocol} & \multirow{2}{*}{Methods} & R@1↑ & R@3↑ & R@5↑ & R@10↑ & MedR↓ & R@1↑ & R@3↑ & R@5↑ & R@10↑ & MedR↓ \\
                          &        & \multicolumn{5}{c|}{Text-motion retrieval} & \multicolumn{5}{c}{Motion-text retrieval} \\
\midrule
\multirow{6}{*}{(a) All} & TEMOS~\cite{petrovich2022temos}    & 7.11 & 17.59 & 24.10 & 35.66 & 24.00 & 11.69 & 20.12 & 26.63 & 36.69 & 26.50 \\
                         & Guo et al.~\cite{guo2022generating}   & 3.37  & 10.84 & 16.87 & 27.71 & 28.00 & 4.94  & 10.72 & 16.14 & 25.30 & 28.50 \\
                         & MotionCLIP~\cite{tevet2022motionclip}  & 4.87 & 14.36 & 20.09 & 31.57 & 26.00 & 6.55  & 17.12 & 25.48 & 34.97 & 23.00 \\
                         & TMR~\cite{petrovich2023tmr}   & 7.23 & 20.36 & 28.31 & 47.00 & 17.00 & 11.20 & 20.12 & 28.07 & 38.55 & 18.00 \\
                         & LAVIMO(3-modal)~\cite{yin2024tri}   & 10.16 & 24.61 & 34.57 & 49.80 & 11.00 & 15.43 & 26.95 & 34.57 & 53.32 & 10.00 \\ 
                         & Ours(3-modal)    & 12.26 & 31.61 & 40.33 & 53.95 & 9.00 & 13.62 & 29.16 & 38.15 & 53.95 & 9.00 \\
                         & Ours(4-modal)     & \textbf{13.44} & \textbf{34.88} & \textbf{44.14} & \textbf{60.22} & \textbf{7.00} & \textbf{16.08} & \textbf{36.24} & \textbf{46.87} & \textbf{58.04} & \textbf{6.00} \\ 
                         \midrule
\multirow{6}{*}{(b) All with threshold} & TEMOS~\cite{petrovich2022temos}    & 18.55 & 30.84 & 42.29 & 56.37 & 7.00  & 17.71 & 28.80 & 35.42 & 47.11 & 13.25 \\
                         & Guo et al.~\cite{guo2022generating}   & 13.25 & 29.76 & 39.04 & 49.52 & 11.00 & 10.48 & 20.48 & 27.95 & 38.55 & 17.25 \\
                         & MotionCLIP~\cite{tevet2022motionclip}  & 13.79 & 31.45 & 49.23 & 53.01 & 9.00  & 13.24 & 29.53 & 38.06 & 50.23 & 10.00 \\
                         & TMR~\cite{petrovich2023tmr}   & 24.58 & 41.93 & 50.48 & 60.36 & 5.00  & 19.64 & 32.53 & 41.20 & 53.01 & 9.50  \\
                         & LAVIMO(3-modal)~\cite{yin2024tri}   & 30.86 & 48.63 & 59.96 & 74.22 & 4.00  & 25.98 & 38.28 & 45.70 & 63.09 & 6.50  \\
                         & Ours(3-modal)             & 32.15 & 56.40 & 67.30 & 79.56 & 3.00  & 32.43 & 53.41 & 62.94 & 73.84 & \textbf{3.00}  \\
                         & Ours(4-modal)             & \textbf{36.51} & \textbf{59.40} & \textbf{70.03} & \textbf{80.38} & \textbf{2.00} & \textbf{33.51} & \textbf{54.77} & \textbf{67.30} & \textbf{79.84} & \textbf{3.00} \\ \midrule
\multirow{6}{*}{(c) Small batches~\cite{guo2022generating}} & TEMOS~\cite{petrovich2022temos}    & 43.88 & 67.00 & 74.00 & 84.75 & 2.06  & 41.88 & 65.62 & 75.25 & 85.75 & 2.25  \\
                         & Guo et al.~\cite{guo2022generating}     & 42.25 & 75.12 & 87.50 & 96.12 & 1.88  & 39.75 & 73.62 & 86.88 & 95.88 & 1.95  \\
                         & MotionCLIP~\cite{tevet2022motionclip}  & 41.29 & 68.50 & 78.83 & 90.55 & 1.73  & 39.55 & 68.13 & 77.94 & 90.85 & 2.16  \\
                         & TMR~\cite{petrovich2023tmr}   & 49.25 & 78.25 & 87.95 & 95.95 & 1.50  & 50.12 &  76.88 & 88.88 & 94.75 & 1.53  \\
                         & LAVIMO(3-modal)~\cite{yin2024tri}   & 58.10 & 86.34 & 93.08 & 96.47 & 1.08  & 60.23 & 77.52 & 93.22 & 95.87 & 1.20  \\ 
                         & Ours(3-modal)             & 62.67 & 88.56 & 94.28 & 97.55 & \textbf{1.01}  & 62.67 & 83.11 & 95.37 & 98.91 & \textbf{1.01}  \\
                         & Ours(4-modal)    & \textbf{66.21} & \textbf{90.46} & \textbf{95.37} & \textbf{99.46} & \textbf{1.01}  & \textbf{64.85} & \textbf{87.47} & \textbf{95.91} & \textbf{99.46} & \textbf{1.01} \\ 
                         \midrule
                         & & \multicolumn{5}{c|}{Video-to-Motion Retrieval} & \multicolumn{5}{c}{Motion-to-Video Retrieval} \\
                         \midrule
\multirow{3}{*}{(a) All} & LAVIMO(3-modal)~\cite{yin2024tri}   & 59.76  & 83.59 & 91.40 & 96.09 & \textbf{1.00} & 63.28 & 85.54 & 90.62 & 96.87 & \textbf{1.00} \\ 
                         & Ours(3-modal)    & 64.85 & 88.82 & 94.00 & 97.82 & \textbf{1.00} & 60.21 & 89.10 & 94.82 & 97.27 & \textbf{1.00} \\
                         & Ours(4-modal)     & \textbf{65.40} & \textbf{89.92} & \textbf{95.64} & \textbf{98.09} & \textbf{1.00} & \textbf{62.94} & \textbf{91.28} & \textbf{96.46} & \textbf{99.46} & \textbf{1.00} \\
                         \midrule
& & \multicolumn{5}{c|}{Audio-to-Motion Retrieval} & \multicolumn{5}{c}{Motion-to-Audio Retrieval} \\
                         \midrule
\multirow{1}{*}{(a) All} & Ours(4-modal)  & 11.98 & 29.97 & 39.50 & 55.04 & 9.00 & 14.98  & 32.69 & 40.05 & 52.86 & 9.00 \\
\bottomrule
\end{tabular}%
}
\label{tab:kit}
\end{table*}

\begin{figure*}[h!]
    \centering
    \includegraphics[width=\textwidth]{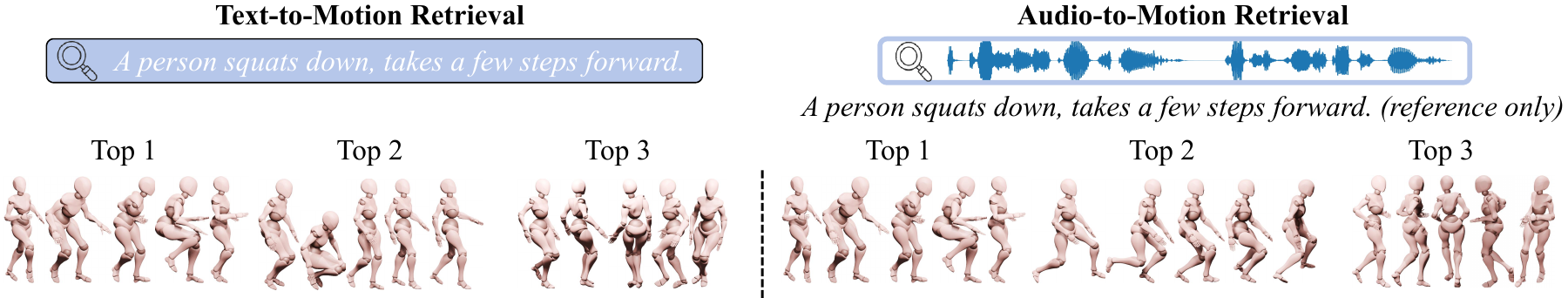}
    \caption{\textbf{Qualitative Comparison of Motion Retrieval Using Text or Audio.} The textual descriptions and audio instructions convey the same meanings. Our approach achieves comparable performance with either modality, highlighting the effectiveness of audio signals as a semantic representation equivalent to text.
    %  on the HumanML3D Dataset
    % Both the Audio-to-motion and Text-to-motion retrieval tasks retrieved the same correct results. Secondly, for the rank2 results, both tasks retrieved similar spinning motions. The same applies to rank3, demonstrating the consistency between speech and text. 
    }
    \label{fig:text-audio_comparison}
\end{figure*}

% At the same time, the 4-modality model exhibits similar performance in the audio-to-motion retrieval task as they do in the text-to-motion retrieval task. We speculate that this may be due to certain consistencies between speech and text, which allow speech to serve as an alternative form of text. Fig.~\ref{fig:text-audio_comparison} shows the similarity of the retrieval results between the audio-to-motion and text-to-motion retrieval tasks. Furthermore, audio provides users with a more convenient and efficient experience in practical applications.
Meanwhile, our 4-modal framework achieves comparable performance in audio-motion retrieval and text-motion retrieval in Tab.~\ref{tab:h3d} and Tab.~\ref{tab:kit}, largely due to inherent consistencies between audio and text, allowing audio to function as an alternative semantic representation. Fig.~\ref{fig:text-audio_comparison} illustrates the comparable qualitative results in motion retrieval using the two modalities. Moreover, audio offers a more convenient and efficient user experience in practical applications.

\begin{table*}[h!]
\centering
\caption{
\textbf{Results on Models Trained on Original and Oral Dataset.} Our method and LAVIMO~\cite{yin2024tri} are evaluated on the Oral dataset. The results reveal a noticeable performance drop when trained on the Original dataset, which we attribute to the stylistic differences between conversational and formal expressions. This highlights the necessity of constructing dedicated oral audio datasets for practical applications.
% This table presents the performance of our model and the LAVIMO~\cite{yin2024tri} model on two audio datasets. 
% The results show that the model trained with oral audio outperforms the one trained with original audio, highlighting the importance of building dedicated oral speech datasets for practical applications.
}
\resizebox{0.9\textwidth}{!}{%
\begin{tabular}{@{}l|l|ccccc|ccccc@{}}
\toprule
\multirow{3}{*}{Methods} & \multirow{3}{*}{Train Set} & R@1↑ & R@3↑ & R@5↑ & R@10↑ & MedR↓ & R@1↑ & R@3↑ & R@5↑ & R@10↑ & MedR↓ \\ \cmidrule{3-12}
                        & & \multicolumn{5}{c|}{Audio-to-Motion Retrieval} & \multicolumn{5}{c}{Motion-to-Audio Retrieval} \\ \midrule
                        \multirow{3}{*}{LAVIMO~\cite{yin2024tri}} &  Oral & 7.78 & 18.07 & 25.81 & 39.06 & 18.00 & 8.95 & 19.45 & 27.27 & 38.81 & 19.00 \\  
                        & Original  & 6.19 & 14.90 & 22.22 & 34.99 & 21.00 & 6.69 & 16.11 & 23.35 & 35.11 & 21.00 \\ 
                        \cmidrule{2-12}
                        & & -1.59 & -3.17 & -3.59 & -4.07 & 3.00 & -2.26 & -3.34 & -3.92 & -3.70 & 2.00 \\
                        \midrule
                        \multirow{3}{*}{Ours} & Oral  & 11.55 & 24.64 & 34.73 & 48.95 & 11.00 & 12.30 & 25.20 & 34.06 & 45.19 & 14.00 \\
                        &  Original  & 7.91 & 20.97 & 30.43 & 44.45 & 13.00 & 9.50 & 21.89 & 29.38 & 42.61 & 15.00 \\
                        \cmidrule{2-12}
                        & & -3.64 & -3.67 & -4.30 & -4.50 & 2.00 & -2.80 & -3.31 & -4.68 & -2.58 & 1.00 \\
\bottomrule
\end{tabular}
}
\label{tab:OralvsOriginal}
\end{table*}

\begin{figure*}[h!]
  \centering
  \includegraphics[width=\textwidth]{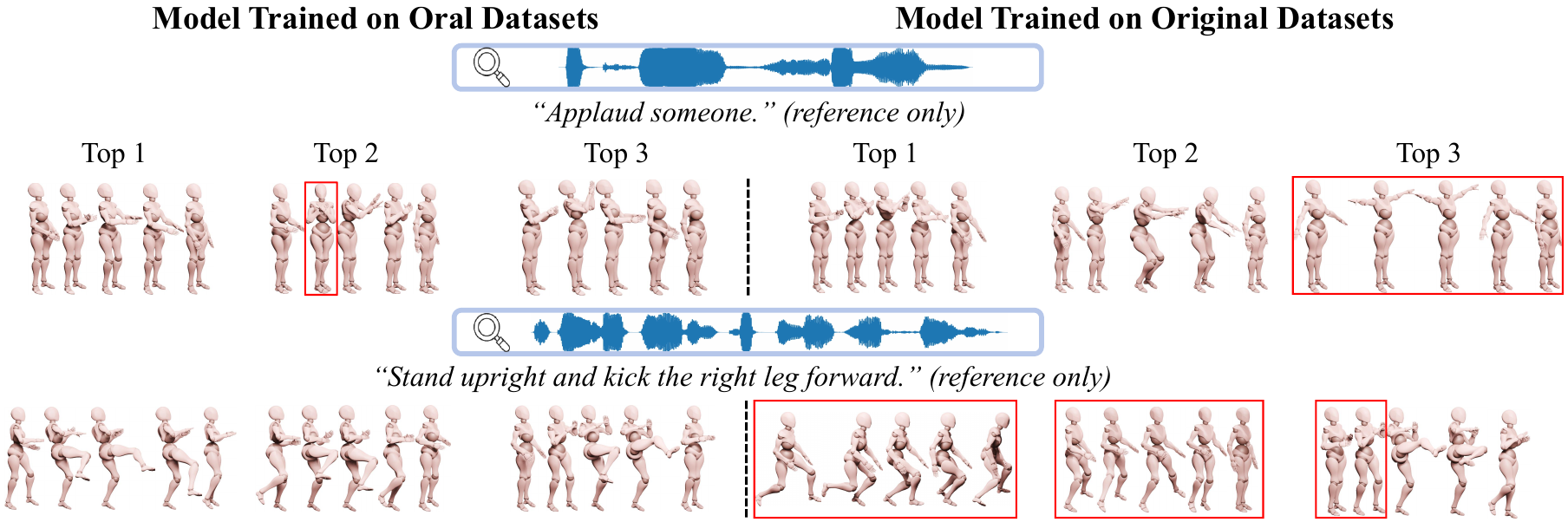}
  \caption{\textbf{Qualitative Comparison of Models Trained on Original and Oral Dataset.} Models trained on the Original dataset struggle to retrieve the correct motions using spoken-style audio signals in Oral dataset, highlighting the importance of our proposed dataset for more conversational applications.
    }
    \label{fig:oral_comparison}
\end{figure*}

\subsection{Results on Oral Dataset}
% As discussed in Sec.~\ref{sec:data-aug}, the textual inputs in the original datasets often differ significantly from the words, sentence structures, and expressions used in users' audio inputs. To assess the impact of this, we train our method and  LAVIMO~\cite{yin2024tri} on either Oral or Original datasets, and test on the Oral test set. Both methods are 3-modal, including motion, text, and audio modalities.

% The experimental results on HumanML3D~\cite{guo2022generating} dataset, shown in Tab.~\ref{tab:OralvsOriginal}. Models trained on the Original datasets perform worse when tested on the Oral datasets, compared to those trained on the Oral datasets. The qualitative results are shown in Fig.~\ref{fig:oral_comparison}. This highlights the necessity of building a dedicated oral datasets to better align with practical applications. This difference arises due to the higher inherent diversity of the oral datasets, which enhances the model's robustness in handling different forms of instructions. 

The textual inputs in the Original dataset often differ significantly from the words, sentence structures, and expressions found in users' audio inputs. To evaluate the impact of this discrepancy, we train both our method and also LAVIMO~\cite{yin2024tri} on either the Oral or Original dataset and test them on the Oral test set. Both models operate in a 3-modal setting, including motion, text, and audio modalities.

The experimental results on the HumanML3D~\cite{guo2022generating} dataset, presented in Tab.~\ref{tab:OralvsOriginal}, show that models trained on the Original dataset perform much worse compared to those trained on the Oral dataset. Qualitative results in Fig.~\ref{fig:oral_comparison} further illustrate this trend. These findings underscore the necessity of constructing dedicated oral datasets to better align with real-world applications. The performance gap stems from the conversational style in oral datasets, which enhances the model’s robustness in handling varied instructions.

% At the same time, the models trained on both the original and oral datasets exhibit similar performance in the audio-to-motion retrieval task as they do in the text-to-motion retrieval task. This supports our hypothesis that audio signals could serve as an alternative to text. Additionally, audio provide users with a more convenient and efficient experience in practical applications.

% \begin{table*}[h]
% \centering
% \caption{
% \textcolor{blue}{\textbf{Ablation Study of Alignment Losses.} Removing alignment losses significantly degrades performance, confirming their importance for cross-modal learning.}
% }
% \resizebox{0.95\textwidth}{!}{%
% \begin{tabular}{@{}l|ccccc|ccccc@{}}
% \toprule
% \multirow{3}{*}{Methods} & R@1↑ & R@3↑ & R@5↑ & R@10↑ & MedR↓ & R@1↑ & R@3↑ & R@5↑ & R@10↑ & MedR↓ \\ \cmidrule{2-11}
%                         & \multicolumn{5}{c|}{Text-to-Motion Retrieval} & \multicolumn{5}{c}{Motion-to-Text Retrieval} \\ \midrule
%                          Ours     & \textbf{9.41}  & \textbf{21.95} & \textbf{30.07} & \textbf{43.83} & \textbf{14.00} & \textbf{10.03}  & \textbf{22.75} & \textbf{30.40} & \textbf{43.74} & \textbf{14.00} \\
%                          \makecell[l]{Ours w/o Alignment Losses}
%                          & 0.00 & 0.12 & 0.20 & 0.54 & 1114.00 & 0.04 & 0.08 & 0.16 & 0.33 & 1196.50 \\
% \bottomrule
% \end{tabular}
% }
% \label{tab:alignment ablation study table}
% \end{table*}

\begin{table*}[h]
\centering
\caption{
\textbf{Ablation Study on Sequence-level Alignment, Body Partition, and Alignment Losses.} All components contribute to retrieval performance, with body partition and alignment losses showing the most significant impact.}
\resizebox{0.95\textwidth}{!}{%
\begin{tabular}{@{}l|ccccc|ccccc@{}}
\toprule
\multirow{3}{*}{Methods} & R@1↑ & R@3↑ & R@5↑ & R@10↑ & MedR↓ & R@1↑ & R@3↑ & R@5↑ & R@10↑ & MedR↓ \\ \cmidrule{2-11}
                        & \multicolumn{5}{c|}{Text-to-Motion Retrieval} & \multicolumn{5}{c}{Motion-to-Text Retrieval} \\ \midrule
                         Ours     & \textbf{9.41}  & \textbf{21.95} & \textbf{30.07} & \textbf{43.83} & \textbf{14.00} & \textbf{10.03}  & \textbf{22.75} & \textbf{30.40} & \textbf{43.74} & \textbf{14.00} \\
                         \makecell[l]{Ours w/o Sequence-level Alignment}
                         & 9.15 & 20.49 & 29.10 & 42.82 & 14.00 & 9.11 & 21.53 & 29.98 & 43.20 & 15.00 \\
                         Ours w/o Body-partition   & 7.77 & 17.69 & 25.63 & 37.51 & 18.00 & 6.98 & 18.02 & 24.55 & 37.26 & 20.00 \\
                         \makecell[l]{Ours w/o Alignment Losses}
                         & 0.00 & 0.12 & 0.20 & 0.54 & 1114.00 & 0.04 & 0.08 & 0.16 & 0.33 & 1196.50 \\  \midrule
                         & \multicolumn{5}{c|}{Video-to-motion Retrieval} & \multicolumn{5}{c}{Motion-to-video Retrieval} \\
                         \midrule
                         Ours     & \textbf{64.78}  & \textbf{88.33} & \textbf{93.56} & \textbf{96.65} & \textbf{1.00} & \textbf{66.54}  & \textbf{88.79} & \textbf{94.48} & \textbf{97.53} & \textbf{1.00} \\ 
                         \makecell[l]{Ours w/o Sequence-level Alignment}        & 50.36 & 75.91 & 84.69 & 93.84 & 1.50 & 51.15 & 78.04 & 86.11 & 93.81 & \textbf{1.00} \\
                         Ours w/o Body-partition     & 34.04 & 61.39 & 73.14 & 83.64 & 2.00 & 40.73 & 68.25 & 79.00 & 89.12 & 2.00 \\
\bottomrule
\end{tabular}
}
\label{tab:ablation study table}
\end{table*}

\begin{table*}[h]
\centering
\caption{\textbf{Retrieval Performance under Varying Sequence Lengths.} Retrieval performance improves with longer sequence lengths, suggesting that richer temporal and semantic information enhances alignment accuracy.} 
\resizebox{0.95\textwidth}{!}{%
\begin{tabular}{@{}c|c|ccccc|ccccc@{}}
\toprule
\multirow{1}{*}{Text} & \multirow{1}{*}{Motion} & \multicolumn{5}{c|}{Text-motion retrieval} & \multicolumn{5}{c}{Motion-text retrieval} \\
 Token Length & Token Length & R@1↑ & R@3↑ & R@5↑ & R@10↑ & MedR↓ & R@1↑ & R@3↑ & R@5↑ & R@10↑ & MedR↓ \\
\midrule
\multirow{6}{*}{}32 & 192 & \textbf{8.78}  & \textbf{19.19} & \textbf{27.10} & \textbf{40.69} & \textbf{16.00} & 9.32  & \textbf{20.78} & \textbf{27.64} & \textbf{41.61} & \textbf{15.00} \\
16 & 96 & 7.36 & 16.10 & 23.84 & 35.63 & 20.00 & 8.74 & 19.41 & 25.44 & 39.10 & 18.00 \\
8 & 48 & 6.98 & 15.89 & 22.71 & 35.59 & 22.50 & 7.94 & 19.02 & 25.01 & 37.35 & 20.00 \\
1 & 1 & 6.37  & 15.60 & 21.95 & 33.67 & 24.00 & \textbf{9.72}  & 18.73 & 25.00 & 36.55 & 22.50 \\ 
\bottomrule
\end{tabular}%
}
\label{tab:variety-length}
\end{table*}

\begin{figure*}[h!]
    \centering
    \includegraphics[width=\textwidth]{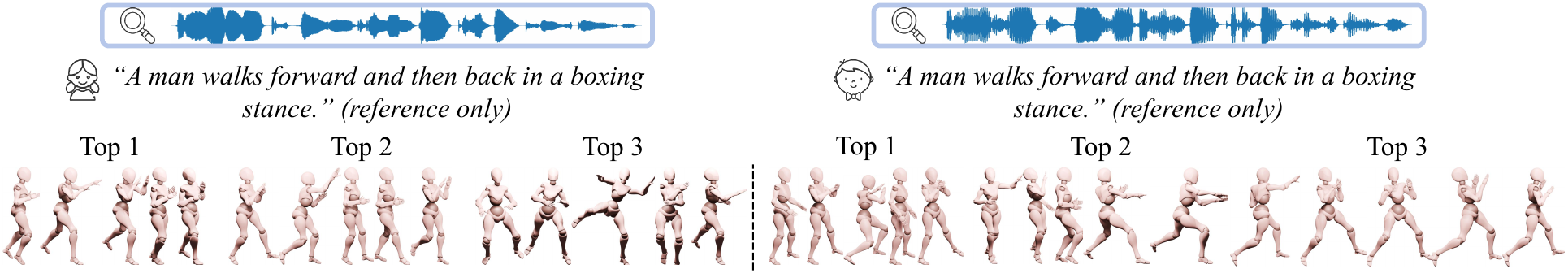}
    \caption{\textbf{Motion Retrieval Using Real Audio Signals.} 
    % We capture real audio signals from three different subjects speaking the same content. The retrieved motions are presented accordingly. The results highlight the generalization ability of our model for real audio motion retrieval.
    We collect real audio recordings from two different subjects, each speaking the same content, and use them for motion retrieval. The results demonstrate that our model, trained on synthetic audio, effectively generalizes to real audio signals, showcasing its robustness in audio-driven motion retrieval.
    }
    \label{fig:audio performance}
\end{figure*}

\subsection{Ablation Study}
In this section, we conduct ablation studies to evaluate the impact of different components on overall performance, as shown in, Table~\ref{tab:ablation study table}, Table~\ref{tab:audio-compression}, and Table~\ref{tab:sim_ablation}.

\textbf{Alignment Losses.} To assess the importance of alignment losses, we conducted ablation studies on the HumanML3D~\cite{guo2022generating} dataset using our 4-modality model. Specifically, we removed all alignment losses and trained the model using only the reconstruction loss. As shown in Tab.~\ref{tab:ablation study table}, this results in a significant performance drop across multiple evaluation metrics, demonstrating that alignment losses are essential for learning meaningful cross-modal representations and ensuring effective retrieval.

\textbf{Sequence-Level Alignment.} Prior methods rely on a learnable global token to summarize the sequence for cross-modal alignment. In contrast, our method directly computes similarities over all token representations, enabling finer sequence-level alignment and capturing richer temporal structure. To verify the effectiveness of our sequence-level alignment framework, we retain all other model structures and parameter settings but replace our method with the global-token-based alignment in the encoders of each modality. As shown in Tab.~\ref{tab:ablation study table}, our fine-grained contrastive learning mechanism significantly improves retrieval performance, highlighting the benefits of modeling token-level correspondence rather than compressing modality information into a single representation. Additionally, we conducted extra ablation study to evaluate the impact of sequence length in motion retrieval, as shown in Tab.~\ref{tab:variety-length}. Overall, retrieval performance improves as the token sequence length increases, indicating that longer token sequences provide more informative representations for cross-modal alignment.

% Prior methods usually rely on global alignment, where modality features are first extracted using a single global token and then aligned with other modality's feature token. To verify the effectiveness of our proposed sequence-level alignment framework, we retain all other model structures and parameter settings but replace sequence-level alignment with a global token in the encoders of each modality. As shown in Tab.~\ref{tab:ablation study table}, our fine-grained contrastive learning mechanism improves retrieval performance by enabling the model to capture richer sequential information.

% Many previous methods commonly use the global token approach, where modality features are first extracted by a global token and then processed for downstream tasks. However, when using sequence-level alignment, our proposed framework significantly outperforms the global token approach in terms of performance. To validate this, we retain all other model structures and parameter settings, only using a global token in the encoders of each modality. The output global token is then passed to the contrastive learning module to compute similarity for feature alignment. Our evaluation, as shown in Table~\ref{tab:ablation study table}, demonstrates that the fine-grained contrastive learning mechanism improves retrieval performance by helping the model capture information in the sequence.

\textbf{Body-Partition Motion Representation.} Previous motion retrieval methods~\cite{yin2024tri} typically overlook the segmentation of human body parts, despite the fact that human motion recognition inherently depends on assessing the state of different body regions. To examine this aspect, we compare our method with a variant that does not perform fine-grained motion segmentation. We keep all other model structures and parameter settings unchanged, processing raw human motion data directly through the transformer in the motion modality encoder before applying downstream tasks. As shown in Tab.~\ref{tab:ablation study table}, our body-partition mechanism enhances model performance by effectively segmenting human motion, allowing it to focus on key body-part-specific information.

\textbf{Comparison of Audio Feature Compression Methods.} The extracted audio features exhibit significant variation in sequence length across different samples, making an effective compression strategy essential for subsequent audio-conditioned motion retrieval.   We explored several methods for compressing these variable-length sequences, including Average Pooling (AvgPool) to fixed lengths (e.g., 4, 2) and 1D convolution (Conv1D), in order to obtain compact and consistent audio representations. The results, presented in Tab.~\ref{tab:audio-compression}, are based on models trained and tested on the Original HumanML3D dataset. The results show that AvgPool yields suboptimal performance, likely due to its lack of learnable parameters, which limits its ability to adapt to diverse audio patterns. Similarly, Conv1D underperforms, as its fixed receptive field fails to capture long-range temporal dependencies. In contrast, our proposed feature compression module effectively handles the large variation in audio sequence lengths and produces more semantically meaningful audio representations, leading to improved performance in audio-to-motion retrieval.

\begin{table*}[h]
\centering
\caption{
\textbf{Ablation Study of Audio Feature Compression Methods.} Our proposed memory-retrieval-based module outperforms alternative architectures by effectively compressing variable-length audio features into compact and informative representations, which are crucial for accurate audio-conditioned motion retrieval.
}
\resizebox{0.9\textwidth}{!}{%
\begin{tabular}{@{}l|ccccc|ccccc@{}}
\toprule
\multirow{3}{*}{Methods} & R@1↑ & R@3↑ & R@5↑ & R@10↑ & MedR↓ & R@1↑ & R@3↑ & R@5↑ & R@10↑ & MedR↓ \\ \cmidrule{2-11}
                        & \multicolumn{5}{c|}{Audio-to-Motion Retrieval} & \multicolumn{5}{c}{Motion-to-Audio Retrieval} \\ \midrule
                         Ours     & \textbf{16.19}  & \textbf{31.91} & \textbf{40.28} & \textbf{53.20} & \textbf{9.00} & \textbf{17.40}  & \textbf{33.12} & \textbf{41.03} & \textbf{51.48} & \textbf{10.00} \\                    
                         \makecell[l]{AvgPool-4}
                         & 9.33 & 22.58 & 31.33 & 45.29 & 13.00 & 12.34 & 26.18 & 34.30 & 45.88 & 13.00 \\
                         \makecell[l]{AvgPool-2}
                         & 10.00 & 22.00 & 30.82 & 45.42 & 13.00 & 11.71 & 25.26 & 34.25 & 46.17 & 13.00 \\
                         \makecell[l]{Conv1D}
                         & 9.03 & 21.62 & 30.11 & 44.79 & 13.00 & 12.25 & 24.34 & 33.25 & 46.38 & 13.00 \\
\bottomrule
\end{tabular}
}
\label{tab:audio-compression}
\end{table*}

% Previous motion retrieval works~\cite{yin2024tri} usually neglect human body partition. However, human motion recognition relies on assessing the state of different body parts. Based on this premise, we compared our method with the one that does not perform fine-grained segmentation of motions. We kept all other model structures and parameter settings unchanged and directly processed the raw human motion data through the transformer in the motion modality encoder before proceeding to downstream tasks. The comparative results, shown in Table~\ref{tab:ablation study table}, reveal that our body-partition mechanism helps the model focus on body part information by effectively segmenting human motion.

\textbf{Token pair in sequence-level alignment.} Our proposed sequence-level representation offers a more expressive and effective alternative to traditional global representations in contrastive learning, which can be seen as a special case where the sequence length is effectively reduced to one. As demonstrated in the ablation study (Tab.~\ref{tab:sim_ablation}), sequence-level alignment consistently outperforms global alignment across most retrieval metrics. Moreover, incorporating global representations with sequence-level similarity yields only marginal improvements, indicating that fine-grained token-level interactions already capture the majority of alignment benefits. To further address concerns regarding potential bias in the max-based sequence-level alignment loss, we compare it with a mean-based aggregation scheme that averages similarities across all token pairs. The results show that the max-based approach consistently achieves better retrieval performance, suggesting that emphasizing the most aligned token pair more effectively captures the dominant semantic correspondence between sequences.

% Previous methods typically rely on a single global token to represent entire sequences, which may overlook finer temporal or spatial details. In contrast, our sequence-level alignment computes similarities between all token pairs, enabling more fine-grained semantic matching across modalities. Ablation results in Tab.~\ref{tab:sim_ablation} show that sequence-level alignment outperforms global alignment in most retrieval metrics. Moreover, incorporating global representations alongside sequence-level similarity brings only limited improvements, indicating that most of the alignment benefit is already captured by token-level interactions. To address concerns about potential bias from selecting the maximum similarity token pair, we compare max-based and mean-based aggregation schemes. The mean-based variant computes the average similarity over all token pairs, rather than focusing on the most similar pair. The results show that max-pooling achieves consistently better retrieval performance, suggesting that identifying the most aligned token pair better captures the dominant semantic correspondence between sequences.

\begin{table*}[h]
\centering
\caption{\textbf{Ablation on similarity strategies for sequence-level alignment.} Sequence-level alignment outperforms global alignment, while adding global features provides only marginal gains.}
\resizebox{\linewidth}{!}{
\begin{tabular}{l|ccccc|ccccc}
\toprule
\multirow{2}{*}{Method} & \multicolumn{5}{c|}{Text-to-Motion Retrieval} & \multicolumn{5}{c}{Motion-to-Text Retrieval} \\
& R@1↑ & R@3↑ & R@5↑ & R@10↑ & MedR↓ & R@1↑ & R@3↑ & R@5↑ & R@10↑ & MedR↓ \\
\midrule
Global Only & 6.37  & 15.60 & 21.95 & 33.67 & 24.00 & 9.72  & 18.73 & 25.00 & 36.55 & 22.50 \\
Sequence-level Alignment (Mean) & 4.51 & 9.82 & 13.46 & 18.06 & 103.00 & 3.17 & 7.36 & 9.45 & 13.75 & 131.00   \\
Sequence-level Alignment (Max) & 8.78  & 19.19 & 27.10 & 40.69 & 16.00 & 9.32  & 20.78 & 27.64 & 41.61 & 15.00 \\
Global + Sequence-level Alignment (Mean) & 6.23 & 14.42 & 20.53 & 31.53 & 27.00 & 5.60 & 10.53 & 12.63 & 20.24 & 67.00 \\
Global + Sequence-level Alignment (Max) & \textbf{9.70} & \textbf{21.95} & \textbf{30.90} & \textbf{44.29}  & \textbf{14.00} & \textbf{11.75} & \textbf{24.59} & \textbf{31.82} & \textbf{45.00} & \textbf{13.00} \\
\bottomrule
\end{tabular}
}
\label{tab:sim_ablation}
\end{table*}

\subsection{Discussion}
% The framework we designed has been demonstrated to be effective in the experiments; however, there are still several issues that warrant further investigation. 

% Firstly, the audio signals synthesized by Tortoise~\cite{betker2023better} from different speakers results in different retrieved motion results, as shown in Fig.~\ref{fig:audio performance}. This may be because the frequency and waveform of the audio signal are also learned as part of the audio features by the model, causing it to consider not only semantic information but also factors such as timbre and intonation when matching actions.
\begin{figure*}[t]
    \centering
    \includegraphics[width=0.95\textwidth]{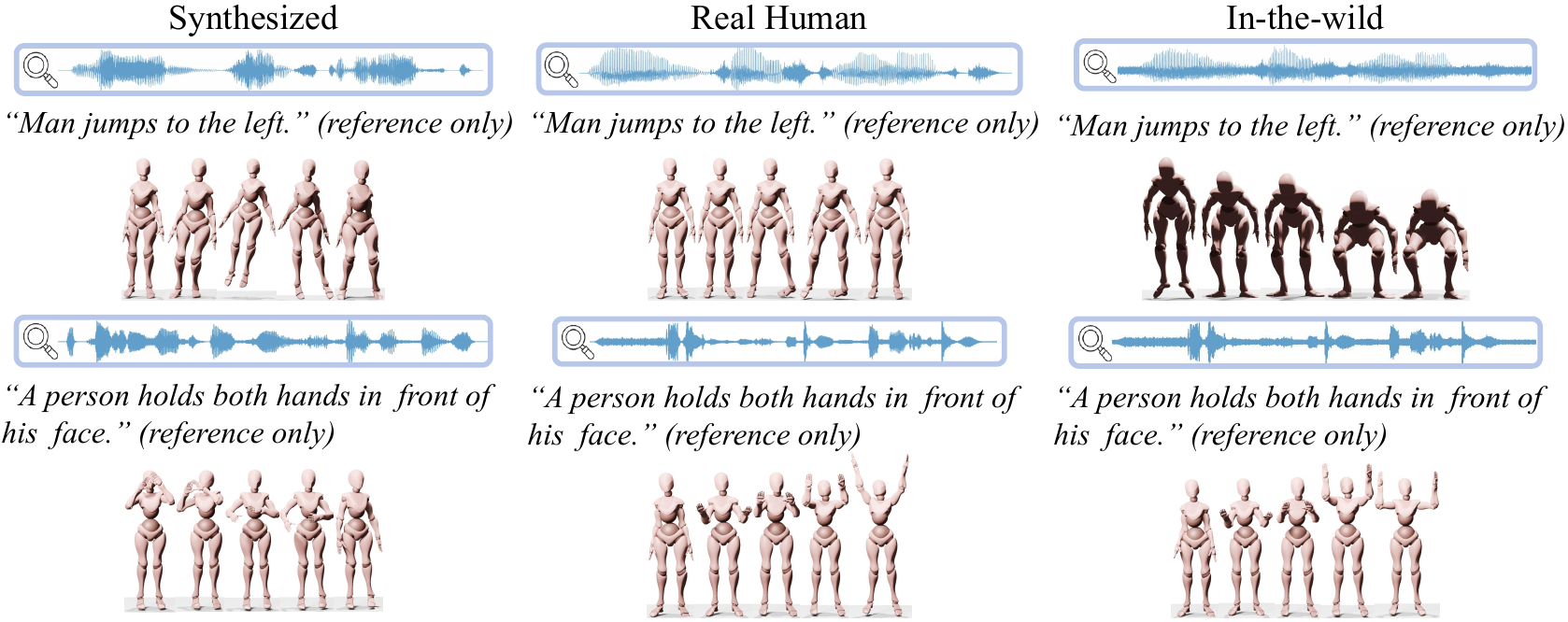}
    \caption{\textbf{Real-world audio-
to-motion retrieval results.} 
    Our model, trained on synthetic audio, generalizes well to some real-world and noisy audios from unseen speakers, showing robust retrieval performance across varied real-world conditions.
    }
    \label{fig:real-world audio}
\end{figure*}

\textbf{Generalization of Synthetic Audio Signals.} The audio signals in our augmented dataset are synthesized using Tortoise~\cite{betker2023better}, which supports diverse speaker voices, enhancing the variety and adaptability of audio across different scenarios. To evaluate the effectiveness of our proposed audio modality data, we collect real audio recordings from two different subjects, each speaking the same content, and use them for motion retrieval. The qualitative results in Fig.~\ref{fig:audio performance} show that our 4-modal model, trained on synthetic audio, successfully generalizes to real audio signals, demonstrating the practicality of our dataset for training models in real-world applications. To further evaluate our model, we conducted real-world audio-to-motion retrieval experiments. Audio signals were recorded from two users in a quiet indoor setting using a microphone with a 12 kHz sampling rate. To simulate in-the-wild conditions, white Gaussian noise was added to the recordings with a signal-to-noise ratio (SNR) of 10 dB. We performed experiments under three audio conditions—synthetic audio, real-world audio, and in-the-wild audio. Visualization results are shown in Fig.~\ref{fig:real-world audio}. Despite being trained on synthetic audio, our model successfully retrieves motions that are semantically aligned with the real users’ inputs, demonstrating strong robustness to real-world audio variations.

% \textcolor{blue}{To evaluate the robustness of the proposed model to variations in speech quality under real-world conditions, we conducted additional experiments using real recorded audio instructions. Specifically, the audio samples were recorded in a quiet indoor environment using a microphone at a sampling rate of 12 kHz. To simulate in-the-wild scenarios, white Gaussian noise was added to the recordings with a signal-to-noise ratio (SNR) of 10 dB. A total of three audio instructions were tested under three conditions: synthetic, real recorded, and noisy inputs. Fig.~\ref{fig:real-world audio} presents one representative example. The retrieval results across the three conditions—synthetic audio, real-world audio, and in-the audio—are not exactly the same, reflecting minor variations in the model's responses to different input qualities. However, the retrieved motions remain semantically consistent with the spoken instructions, showing the robustness of our method to real-world variability in speech input. }

\begin{figure}[t]
    \centering
    \includegraphics[width=0.45\textwidth]{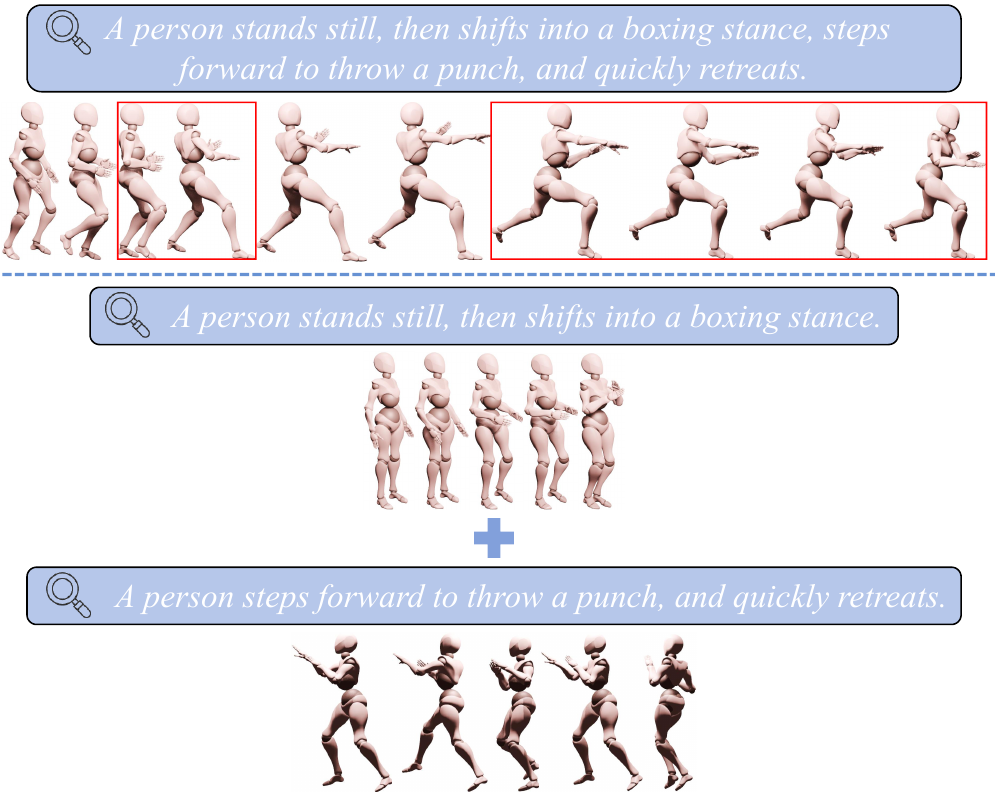}
    \caption{\textbf{Failure Case Analysis.} 
    The retrieval fails when using a long text sequence for retrieval, likely due to the motion length limit set in our model. However, when we split the sentence into two shorter ones, the retrieved motions are correct, where post-processing can be applied to seamlessly stitch them together.
    }
    \label{fig:failure case}
\end{figure}

\textbf{Performance Discrepancy Across Datasets.} We observe that the model achieves better performance on text-motion and video-motion retrieval tasks in the KIT-ML~\cite{plappert2016kit} dataset compared to the HumanML3D~\cite{guo2022generating} dataset. This discrepancy may stem from the smaller scale of KIT-ML, which makes the model more susceptible to overfitting. In contrast, the larger and more diverse HumanML3D dataset presents greater variability and complexity, requiring the model to learn more robust features. Additionally, we find that the performance improvement on HumanML3D is significantly greater than that on KIT-ML for both retrieval tasks. This suggests that the larger dataset provides a more rigorous evaluation of the model’s generalization ability, further validating the effectiveness of our approach.

\textbf{Efficiency Analysis.} Our model is trained on multi-modal inputs to learn a joint embedding space across different modalities, but supports flexible inference with a single modality—such as text-motion or video-motion retrieval. The multi-modal motion retrieval framework incorporates new modules, yet the inference process remains efficient and lightweight. To further assess the efficiency of our model, we provide quantitative latency measurements for both training and inference. For text-motion retrieval, the average inference latency is approximately 92.6 and 90.3 milliseconds per sample on HumanML3D and KIT-ML, respectively, measured over the full test set using a single NVIDIA A6000 GPU. For audio-motion retrieval, the inference latency is 49 and 38 milliseconds per sample on the two datasets. Notably, the average length of text descriptions is approximately 12 for HumanML3D and 8 for KIT-ML, while the average audio durations are around 4.5 and 3.6 seconds, respectively. These results highlight the real-time inference capability of our motion retrieval framework. Additionally, training of the 4-modality model is conducted using four NVIDIA A6000 GPUs, requiring approximately 0.253 hours per epoch (5.96 hours total) on KIT-ML and 0.47 hours per epoch (24.2 hours total) on HumanML3D. In our current setting, the input sequences are relatively short (typically under 200 tokens or frames), so the inference cost is acceptable. For long sequences that may introduce significant latency, we suggest potential solutions such as using chunked computation or gradient accumulation to reduce peak memory usage without compromising training quality, and incorporating sparse attention mechanisms or top-k similarity filtering to avoid unnecessary computation on padded or low-relevance tokens. As discussed, our model supports single-modality inference, maintaining efficient deployment. Despite the introduction of new modules, the inference process remains efficient. Like many multi-modal retrieval frameworks, our method involves a trade-off between performance and computational cost; however, the improved alignment performance justifies the modest increase in training complexity.

\textbf{Failure Case Analysis.} 
% It can be observed in Fig.~\ref{fig:failure case} that when a long textual description is applied, the retrieved top-1 motion is not correct. This is largely due to the motion sequence length limit of 196 in 20 FPS set in our model, which also follows the setting in prior work~\cite{guo2022generating,yin2024tri}. The potential solution to this issue is to divide long sentence into short ones and then retrieve corresponding motion sequences separately. Finally, we can stitch them by aligning and interpolating the global rotation and translation of root joints. 
As shown in Fig.~\ref{fig:failure case}, when a long textual description is used, the retrieved Top-1 motion is incorrect. This issue primarily arises from the motion sequence length limit of 196 at 20 FPS set in our model, following the configuration in prior work\cite{guo2022generating,yin2024tri}. A potential solution is to divide the long sentence into shorter segments and retrieve the corresponding motion sequences separately. These retrieved clips can subsequently be stitched together using motion smoothing techniques or transition-aware interpolation methods to ensure temporal coherence and natural motion flow. 
Another potential approach is to reduce the motion frame rate (FPS) during preprocessing, which allows longer-duration motions to fit within the frame limit imposed by the model.  While this reduces temporal resolution, it enables the model to capture the overall structure and semantics of extended motions more effectively. Finally, incorporating efficient Transformer architectures such as sparse attention or sliding-window mechanisms can enable the model to handle longer motion sequences directly, preserving long-range dependencies without excessive computational cost.

\section{Conclusion}
In this work, we propose a framework that aligns four modalities—text, audio, video, and motion—within a fine-grained joint embedding space, introducing audio for the first time in motion retrieval. To enhance existing text-motion datasets, we synthesize audio from text and incorporate a more conversational, spoken language style to increase audio diversity. Our fine-grained contrastive learning approach aligns modality features at the sequential level, capturing richer details for improved retrieval performance. Experimental results demonstrate that our framework achieves significant improvements in multi-modal motion retrieval tasks, particularly in text-motion, video-motion, and audio-motion retrieval. These advancements provide valuable insights for multi-modal systems in practical applications.

\section*{Acknowledgment}
This work was partially supported by the grants from National Natural Science Foundation of China (62372441), in part by Guangdong Basic and Applied Basic Research Foundation (2023A1515030268), and in part by Shenzhen Science and Technology Program (Grant No. RCYX20231211090127030, JSGG20220831105002004).

% if have a single appendix:
%\appendix[Proof of the Zonklar Equations]
% or
%\appendix  % for no appendix heading
% do not use \section anymore after \appendix, only \section*
% is possibly needed

% use appendices with more than one appendix
% then use \section to start each appendix
% you must declare a \section before using any
% \subsection or using \label (\appendices by itself
% starts a section numbered zero.)
%

% \appendices
% you can choose not to have a title for an appendix
% if you want by leaving the argument blank
% \section{}
% Appendix two text goes here.

% Can use something like this to put references on a page
% by themselves when using endfloat and the captionsoff option.
\ifCLASSOPTIONcaptionsoff
  \newpage
\fi

% trigger a \newpage just before the given reference
% number - used to balance the columns on the last page
% adjust value as needed - may need to be readjusted if
% the document is modified later
%\IEEEtriggeratref{8}
% The "triggered" command can be changed if desired:
%\IEEEtriggercmd{\enlargethispage{-5in}}

% references section

% can use a bibliography generated by BibTeX as a .bbl file
% BibTeX documentation can be easily obtained at:
% http://mirror.ctan.org/biblio/bibtex/contrib/doc/
% The IEEEtran BibTeX style support page is at:
% http://www.michaelshell.org/tex/ieeetran/bibtex/
\bibliographystyle{IEEEtran}
% argument is your BibTeX string definitions and bibliography database(s)
\bibliography{IEEEabrv}
\end{document}